\documentclass{article}
\usepackage{graphicx}
\usepackage{arxiv}

\usepackage[utf8]{inputenc} 
\usepackage[T1]{fontenc}    
\usepackage{hyperref}       
\usepackage{url}            
\usepackage{booktabs}       
\usepackage{amsfonts}       
\usepackage{nicefrac}       
\usepackage{microtype}      
\usepackage{lipsum}		
\usepackage{natbib}
\usepackage{doi}
\usepackage{amsmath}
\usepackage{colortbl}
\usepackage{xcolor} 
\definecolor{Gray}{gray}{0.9} 
\usepackage{float}
\NewDocumentCommand{\rot}{O{60} O{1em} m}{\makebox[#2][l]{\rotatebox{#1}{#3}}}%
\usepackage{tcolorbox}
\usepackage{listings}
\usepackage{epstopdf}
\usepackage{adjustbox}
\usepackage{multirow}
\usepackage{tcolorbox}

\title{Recent Advances in Named Entity Recognition: A Comprehensive Survey and Comparative Study}

\date{} 					

\author{{Imed Keraghel} \\
	Centre Borelli UMR9010\\
 	Université Paris Cité\\
	Paris, France \\
	\texttt{imed.keraghel@u-paris.fr} \\
	\And
	{Stanislas Morbieu} \\
	Kernix Software\\
	Paris, France \\
	\texttt{smorbieu@kernix.com} \\
        \And
        {Mohamed Nadif} \\
	Centre Borelli UMR9010\\
 	Université Paris Cité\\
	Paris, France \\
	\texttt{mohamed.nadif@u-paris.fr} \\
}

\renewcommand{\shorttitle}{Recent Advances in Named Entity Recognition: A Comprehensive Survey and Comparative Study}

\hypersetup{
pdftitle={Recent Advances in Named Entity Recognition: A Comprehensive Survey and Comparative Study},
pdfauthor={Imed ~Hippocampus, Stanislas ~Morbieu, Mohamed ~Nadif},
pdfkeywords={Named Entity Recognition, Information
Extraction, Natural Language Processing, Large Language Models, Machine Learning}
}

\begin{document}
\maketitle

\begin{abstract}
Named Entity Recognition seeks to extract substrings within a text that name real-world objects and to determine their type (for example, whether they refer to persons or organizations). In this survey, we first present an overview of recent popular approaches, including advancements in Transformer-based methods and Large Language Models (LLMs) that have not had much coverage in other surveys. In addition, we discuss reinforcement learning and graph-based approaches, highlighting their role in enhancing NER performance. Second, we focus on methods designed for datasets with scarce annotations. Third, we evaluate the performance of the main NER implementations on a variety of datasets with differing characteristics (as regards their domain, their size, and their number of classes). We thus provide a deep comparison of algorithms that have never been considered together. Our experiments shed some light on how the characteristics of datasets affect the behavior of the methods we compare.
\end{abstract}


\section{Introduction}

Named Entity Recognition (NER) is a subfield of computer science and Natural Language Processing (NLP) that focuses on identifying and classifying entities in unstructured text into predefined categories, such as persons, geographical locations, and organizations \citep{grishman1996message}. Over time, NER has expanded its scope beyond proper names to include more complex concepts \citep{mehmood2023use}, particularly in specialized domains such as biomedicine. For example, in the biomedical field, NER techniques are employed to identify entities such as genes, proteins, and diseases \citep{mesbah2018tse, luo2023aioner}. Consequently, NER has become a crucial component in various modern applications, including machine translation \citep{babych2003improving}, question-answering systems (QA) \citep{molla2006named}, and information retrieval (IR) \citep{guo2009named}.

Historically, early NER systems relied on rule-based approaches with hand-crafted rules, lexicons, and spelling features \citep{rau1991extracting, mikheev1999named, farmakiotou2000rule}. These methods, while simple and interpretable, lacked flexibility and scalability. The introduction of machine learning techniques marked a significant shift in the field, allowing more adaptable and data-driven approaches \citep{chieu2003named, nadeau2007survey, konkol2013crf, shaalan2014survey, eltyeb2014chemical}. With the rise of neural networks, NER systems further improved, particularly with the adoption of deep learning methods, which enabled more sophisticated models capable of capturing complex patterns in text \citep{collobert2011deep, zhao2016ml, zhang2018chinese}. Most recently, Transformer-based architectures have set new standards in NER performance, leading to breakthroughs in the field \citep{labusch2019bert, jeong2022scideberta, vacareanu2024active,shi2024bert}.

\begin{figure}[ht!]
  \centering
  \includegraphics[width=4.99in]{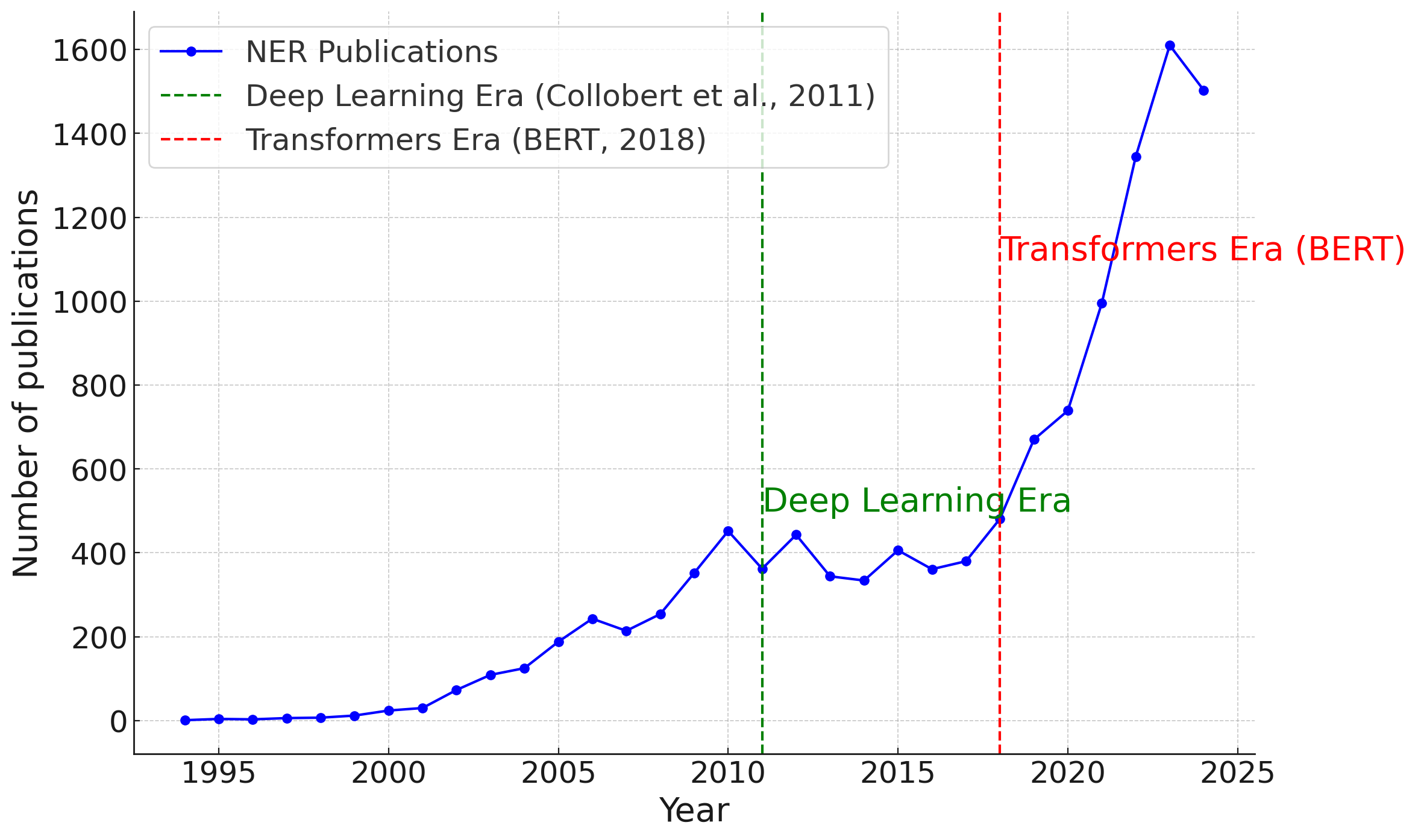}
  \caption{Growth of NER publications.}
  \label{fig.NER.growth}
\end{figure}

This progress is reflected in the substantial growth of NER research publications in the last three decades. As shown in Figure \ref{fig.NER.growth}, the number of NER-related publications in the ACM Computing Surveys database has grown steadily. In the mid-1990s, NER publications were relatively few, reflecting the field’s early focus on rule-based systems. The introduction of statistical models such as hidden Markov models \citep{baum1970maximization} and conditional random fields \citep{lafferty2001conditional} around the year 2000 brought a surge in academic interest. More recently, the period 2018-2024 saw an explosion in NER publications, driven by the adoption of Transformer-based models that improved the performance of NER systems.

To track the evolution of NER and provide structured overviews of recent advances, several surveys have been published documenting the progression from rule-based methods to modern machine learning approaches \citep{nadeau2007survey, shaalan2014survey, goulart2011systematic, marrero2013named}. For example, \citet{goulart2011systematic} specifically reviewed advances in biomedical NER between 2007 and 2009, while \citet{marrero2013named} provided a broader theoretical and practical review of NER techniques. However, many of these early surveys predate the widespread adoption of deep learning and Transformer-based models. More recent surveys have shifted their focus to these modern approaches. For example, several reviews have concentrated on deep learning and Transformer-based methods \citep{pakhale2023comprehensive, jehangir2023survey, li2020survey,10.1145/3445965,10.1145/3594315.3594359,li2022review}. Despite their depth, these surveys often overlook recent advances, such as the integration of Large Language Models (LLMs) and graph-based techniques. Research using graph-based methods, such as the work by \citet{wang2022nested}, focuses on specific challenges such as nested entities but does not address flat-named entities. Similarly, biomedical NER surveys, such as \citet{10.1145/3611651}, mention the use of pre-trained language models, but lack detailed analysis of their application in this domain.

A detailed review of NER in historical documentation by \citet{ehrmann2023named} examines the distinct challenges and strategies relevant to this field. This research provides significant insights into NER applied to historical texts, which often present unique issues such as language evolution, non-standardization, and inconsistent spelling. However, while it addresses these specific challenges in historical documents, the review does not comprehensively cover broader advancements in NER technologies, such as LLMs or graph-based approaches, nor does it extensively discuss strategies to manage limited annotations. Moreover, several recent works that use LLMs for NER have not yet been covered in existing surveys. For example, \citet{wang2023gptner} improve few-shot NER in new domains by incorporating type-specific features, while \citet{ashok2023promptner} uses entity type definitions to enhance few-shot learning.

Another notable gap in the literature is the limited attention paid to methods that address low-resource settings, where annotated data is scarce. Producing annotated datasets is often expensive and time consuming, making it essential to develop methods that can perform effectively with limited data. To the best of our knowledge, no comprehensive survey has yet focused on techniques designed for datasets with scarce annotations. Furthermore, recent work has shown that Reinforcement Learning (RL) can help address the challenge of improving model performance in NER \citep{yang2023gaussian, 10.1007/978-3-030-60450-9_27}. However, this area remains underexplored in existing surveys, and only \citet{pakhale2023comprehensive} briefly mentions the potential of RL in NER.

In this paper, our aim is to address these gaps by providing an all-encompassing review of NER techniques, from early rule-based approaches to the most recent methods, including those relying on LLMs and graph-based approaches. Our study also examines other learning paradigms such as RL. In addition, we focus on methods designed for datasets with scarce annotations and compare NER implementations across various datasets.

Our paper is structured as follows. We begin by defining the task of NER and explaining the different types of named entities. Next, in Section 4, we illustrate some NER applications. Significant methods are discussed in Section \ref{methods}, with an emphasis on LLM, RL, and graph-based approaches. Section \ref{low.resource} covers techniques suitable for scenarios with limited annotated data. Section \ref{frameworks} reviews well-known tools for pre-trained models. Following the explanation of NER evaluation schemes in Section \ref{evaluation}, we provide a variety of useful corpora to the research community in Section \ref{datasets}. For comparative analysis, Section \ref{sec:experiments} includes the application of the latest versions of five popular frameworks on selected datasets. The paper concludes with our findings and future perspectives in Section \ref{conclusion}.

\section{Task definition}\label{ner.task}

NER is a specific task within NLP that involves identifying and categorizing named entities in a text corpus. These entities, defined as words or phrases, refer to real-world objects such as people, organizations, locations, temporal expressions, numerical values, and gene or protein identifiers in the biomedical field \citep{lee2004biomedical,luo2023aioner,wu2024named}. The main goal of NER is to detect and classify entities into predefined semantic categories.

\begin{figure}[ht!]
  \centering
  \includegraphics[width=3.2in]{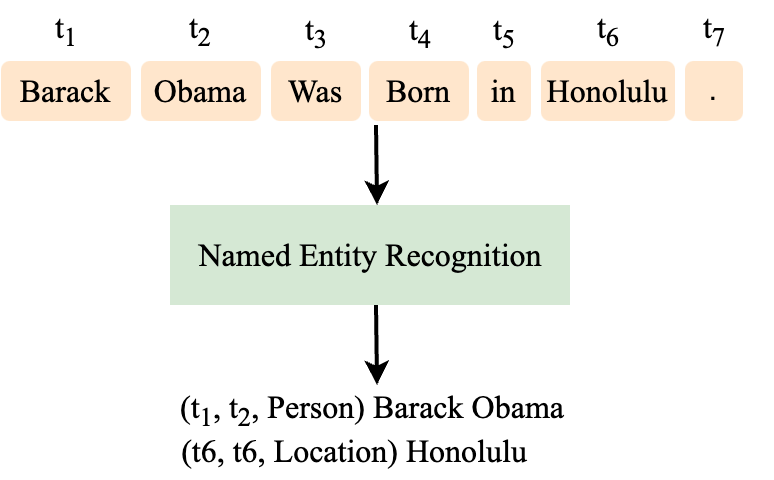}
  \caption{Given a sequence of tokens, NER outputs the boundaries of the named entities along with their associated category.}
  \label{fig.NER.task}
\end{figure}

In a formal context, consider $T$ a sequence of $N$ tokens represented by $T = (t_1, t_2, \dots, t_N)$. NER entails generating a set of tuples $(I_s, I_e, \ell)$, where $s$ and $e$ are integers confined to the interval $[1, N]$. Here, $I_s$ and $I_e$ denote respectively the start and end indices of mentions of a named entity, and $\ell$ indicates the category, from among a set of predefined categories, to which the entity belongs. For example, in the sentence ``Barack Obama was born in Honolulu.'', NER would identify ``Barack Obama'' as a person's name and ``Honolulu'' as a location, as shown in Figure~\ref{fig.NER.task}.

\section{Types of named entities}\label{ner.types}

Named entities are categorized by their structural and contextual features. The three main types are nested, non-continuous, and continuous named entities.

\subsection{Nested named entities}

Nested named entities are entities contained within other entities. For example, in the sentence "Barack Obama was born in Honolulu, Hawaii." the term "Honolulu, Hawaii" is a location entity nested within another (Honolulu being the capital of the island state of Hawaii). Identifying such nested entities is crucial in specialized fields such as biomedical text mining, where entities frequently overlap and are intricately embedded within one another. The management of nested entities generally involves hierarchical models or multi-level tagging methods \citep{alex2007recognising,shibuya2020nested,wang2022nested, shen2021locate}.

\subsection{Non-continued named entities}

Non-continued named entities take the form of singular, contiguous spans within the text. These entities constitute the most elementary form of named entities, each distinctly delineated and separable from others. For example, in the sentence "Google was founded by Larry Page and Sergey Brin." the entities "Google," "Larry Page," and "Sergey Brin" are non-continued named entities. Conventional NER systems are typically able to process such entities using sequence-labeled methods \citep{collobert2011deep,liu2021ner,wang2023gptner}.

\subsection{Continued named entities}

Continued named entities are entities that span multiple, non-contiguous parts of the text. This can occur in cases where the entity is interrupted by other text, but still refers to the same real-world object. An example can be found in the sentence "The patient exhibited a productive cough with white or bloody sputum". Here, "cough white sputum" and "cough bloody sputum" are parts of the same symptom but are separated by other descriptive text. Recognizing continued entities requires models to understand context beyond immediate word sequences, often leveraging more advanced techniques such as attention mechanisms in Transformers \citep{pakhale2023comprehensive}.

\section{Applications of NER}
In this section, we present several illustrative examples of applications for NER.
\begin{itemize}
    \item \textbf{Healthcare and clinical research}: NER is widely used in healthcare for extracting patient-related information from clinical notes, medical literature, and electronic health records (EHRs). By accurately identifying names of drugs, symptoms, diseases, and treatments, NER facilitates the aggregation of critical clinical data, which is essential for patient care and medical research. \citet{liu2023detection} highlighted the role of NER in improving EHR data extraction, and \citet{jagannatha2019overview} introduced the first NLP challenge for extracting medication, indication, and adverse drug events from EHRs. NER can be used to pseudonymize clinical documents, ensuring that patient privacy is maintained while enabling the use of clinical data for research. A comprehensive approach combining rules with deep learning has been developed to address the challenges of de-identification in clinical data warehouses \citep{tannier2024development}.
    \item \textbf{Information extraction}: NER can be used to extract structured data from unstructured text, for example in retrieving the names of persons, organizations, or medical concepts. Studies such as \citep{nadeau2007survey, weston2019named} have explored different ways of improving NER systems. To this end, \citet{etzioni2005unsupervised} examined the use of unsupervised learning approaches, while \citet{weston2019named} investigated the use of deep learning to increase the accuracy and efficiency of entity recognition. These studies are examples of the progress currently being made in increasing the precision of information extraction.
    \item  \textbf{Information retrieval}: NER can significantly improve both traditional and conversational information retrieval (IR) by accurately identifying and classifying entities in user queries and responses. This improves the precision of search results and allows systems to better understand user intent. In IR, improvements come from precise identification of pertinent entities in both search queries and the resulting data \citep{banerjee2019information,cheng2021end}. Research by \citet{cowie1996information} and \citet{etzioni2005unsupervised} confirms that NER improves IR systems. In conversational systems, NER enables accurate understanding of context and user intent. For example, when a user inquires about the weather at a specific location, NER extracts the location entity, allowing the conversational assistant to provide an appropriate response. Studies such as \citep{jayarao2018exploring,cheng2019evaluating, park2023admit} highlight the effectiveness of NER in improving the performance of virtual assistants-.
    \item \textbf{Document summarization}: Integrating NER into the process of document summarization significantly enhances the quality and pertinence of the resulting summaries. By accurately identifying and classifying key entities, NER ensures that the summary encapsulates crucial information regarding these entities. Prior research, such as \citep{khademi2020persian, liu2022summarization, roha2023moo}, underscores the important role of NER in refining document summarization methods.
    \item \textbf{Social media monitoring}: Where it is integrated with social media monitoring, NER enables businesses to automatically identify and categorize mentions of entities like brands and persons. This can help track brand visibility, sentiment analysis, competitor insights, and crisis management. NER is also useful for spotting trends, assessing campaign effectiveness, and leveraging influencer marketing \citep{sufi2022tracking}. Further studies by \citet{ji2024cmner} emphasize the value of NER in processing and analyzing social media data.
    \item \textbf{Named entity disambiguation}: NER can help disambiguate entities with the same term \citep{al2016arabic}. For example, "Apple" can refer to the company or the fruit, and NER can determine the correct interpretation based on the surrounding context. Studies by \citet{bunescu2006using} and \citet{dredze2010entity} further explore techniques for improving named entity disambiguation.
    \item \textbf{Question answering}: NER can play a role in answering questions that require pinpointing particular entities  \citep{molla2006named}. For example, in a question like "When did Steve Jobs die?", NER can identify "Steve Jobs" as a person and extract the death date. Works by \citet{molla2006named} and \citet{zhang2016question} demonstrate the value of NER in improving question answering systems.
    \item \textbf{Language translation}: NER can help improve the accuracy of machine translation by preserving named entities \citep{li2020language}. Additional work by \citet{hkiri2017arabic} supports the role of NER in refining language translation processes.
\end{itemize}
\section{Methods} \label{methods}
\begin{figure}
  \centering
  \includegraphics[width=13cm]{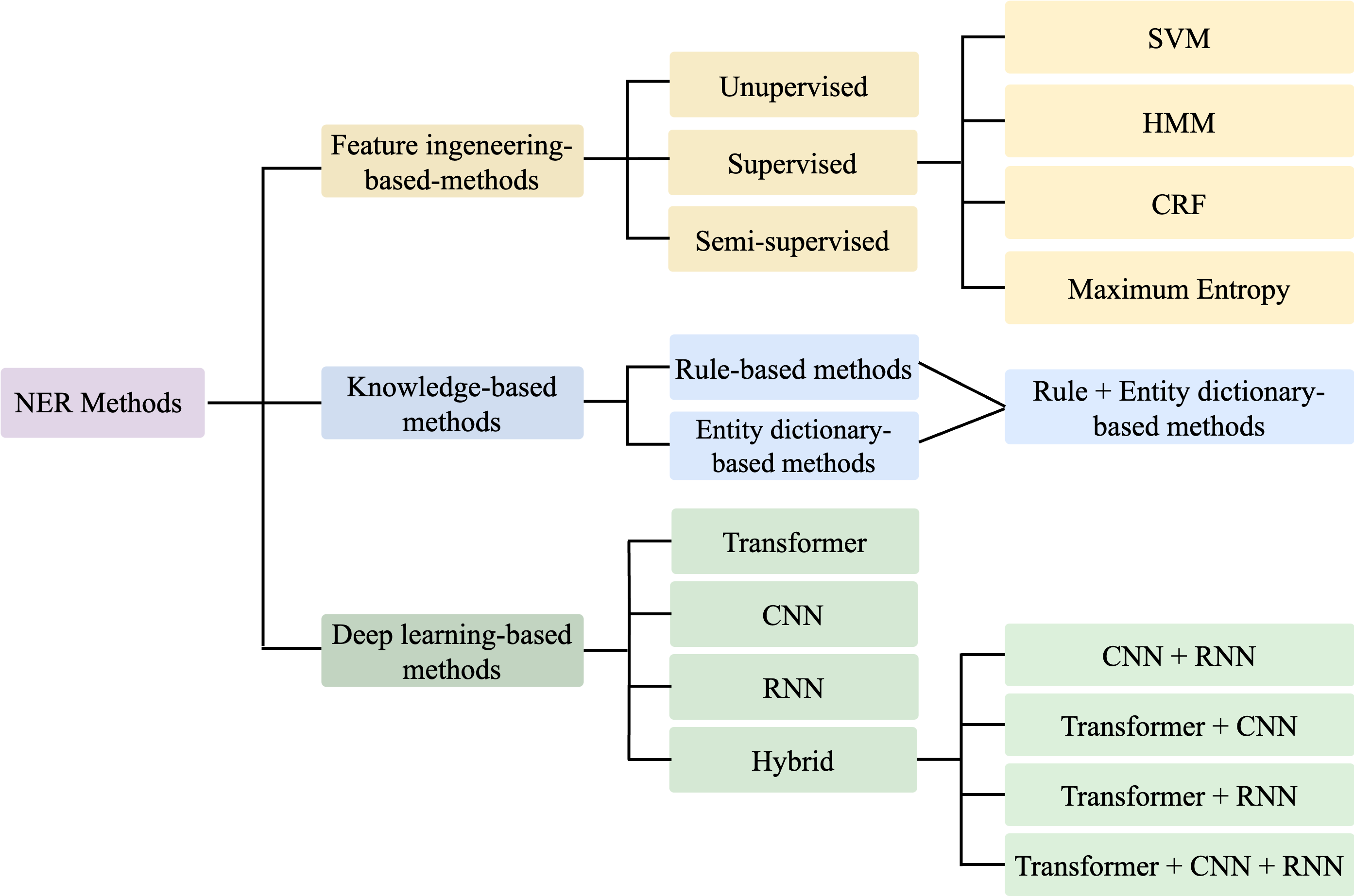}
  \caption{Main approaches to NER.}
  \label{fig.main.app}
\end{figure}

In this section, we explore the various approaches used for NER. These methodologies span a broad range of techniques, from knowledge-based systems to modern deep learning architectures. An overview of these approaches is illustrated in Figure~\ref{fig.main.app}.
\subsection{Knowledge-based methods} \label{methods.knowledge}

Knowledge-based methods have been a foundational part of NER, particularly in the early stages of the field. These methods, which originate from linguistic principles, rely on predefined rules and lexical resources to identify named entities. For example, \citet{borkowski1966system} presented an algorithm that uses rule-based lists and lexical markers, such as capitalization patterns, to identify company names. This rule-driven approach allows systems to detect entities based on consistent patterns, such as prefixes like "Mr." or "Ms." that signal the presence of a named entity. Another notable example is the \emph{CasEN} transducer cascade \citep{maurel2011cascades}, designed specifically to recognize French-named entities.

\begin{figure}[ht!]
  \centering
  \includegraphics[width=3.5in]{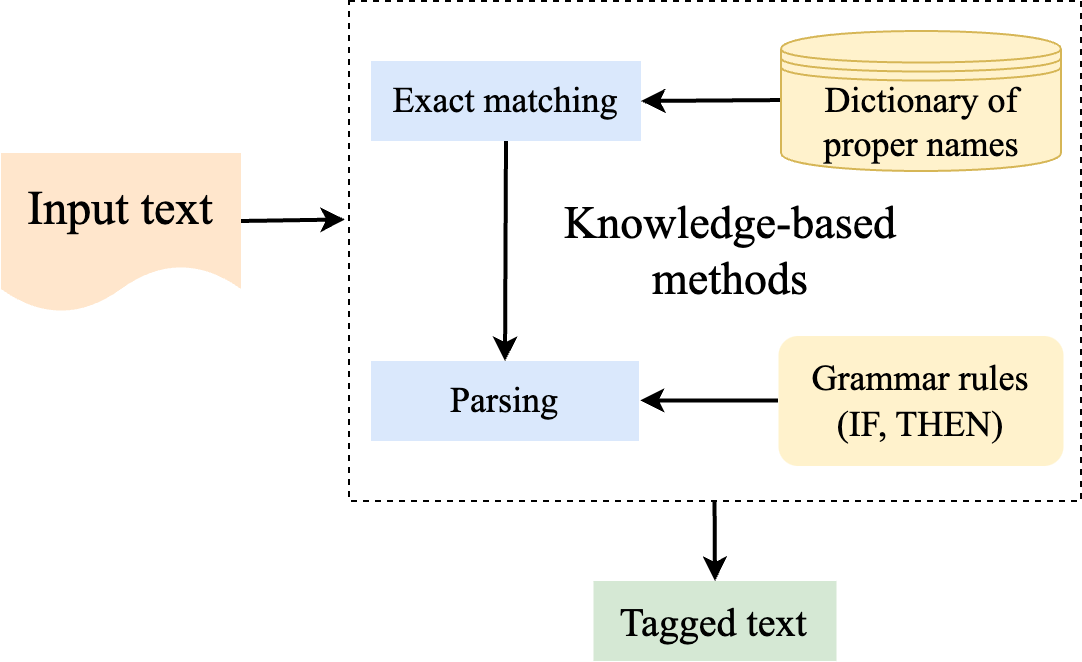}
  \caption{The architecture of knowledge-based methods for NER.}
  \label{fig.knowledge}
\end{figure}

These methods, which do not require annotated data, are primarily based on rules and gazetteers. Gazetteers serve as essential resources, providing a collection of domain-specific entities that improve recognition performance. Figure \ref{fig.knowledge} shows a typical architecture used in knowledge-based NER systems. Architectures generally involve three key components: (1) a set of rules for identifying named entities, (2) optional gazetteers for additional context, and (3) an extraction engine that applies the rules and gazetteers to the input text. Despite the effort required to build these systems ``manually'', they offer robust performance in well-defined domains \citep{btoush2016rule,eftimov2017rule} with well-adapted gazetteers \citep{hanisch2005prominer,sekine2004definition}.

One of the key applications of knowledge-based methods is in biomedical NER, where systems such as \textit{ProMiner} \citep{hanisch2005prominer} have been developed to identify gene and protein names in texts. These systems use synonym gazetteers and specialized detection procedures to address challenges such as synonym ambiguity and case sensitivity. Similarly, \citet{quimbaya2016named} applied a gazetteer-based technique to electronic health records, obtaining an improvement in recall with minimal impact on precision (see Section \ref{metrics}).

Other studies have applied knowledge-based methods in a variety of domains. For example, a rule-based approach was used to extract dietary recommendations from unstructured text, demonstrating the flexibility and domain-specific applicability of such methods \citep{eftimov2017rule}. Hybrid methods that combine rule-based approaches with machine learning have been shown to improve the accuracy of entity recognition in complex texts such as biomedical literature \citep{zhang2013unsupervised}. A comprehensive approach that combined rules and deep learning techniques for pseudonymizing clinical documents \citep{tannier2024development} serves as a good illustration of the importance and challenges of deidentification in clinical data warehouses.

An interesting extension to knowledge-based approaches involves the use of factored sequence labeling to extract methodology components from AI research articles. \citet{ghosh2023extracting} propose a data-driven factored sequence labeling approach that leverages both ontology-based and data-driven techniques. The ontology-based technique uses predefined categories from knowledge bases such as PaperswithCode, while the data-driven variant employs clustering on sentence embeddings to dynamically identify emerging methodologies. This combined approach allows for more precise extraction of scientific concepts, particularly in dynamic fields where terminology evolves rapidly. The factorized method captures the dependencies between methodology names and their contexts, thus overcoming the limitations of traditional rule-based approaches, especially for newly introduced concepts.

For this family of methods, \emph{precision} is typically high, while \emph{recall} tends to be low due to domain- or language-specific rules and incomplete gazetteers. Moreover, the process of introducing new rules and gazetteers is costly. These methods, while resource-intensive, provide robust performance in well-defined domains and continue to be an essential component of NER systems. They remain relevant today, despite having been around for some time, and recent studies continue to use them, as indicated by \citep{wu2022rule,mengliev2023developing}.

\subsection{Feature-engineering-based methods} \label{methods.machine.learning}

As NER evolved, the need to automate the extraction of named entities led to the development of methods based on feature engineering. The idea behind these approaches was to reduce manual rule-setting by focusing on identifying key features that could be used for NER tasks within the text. Feature-engineering-based methods can be broadly categorized into unsupervised, supervised, and semi-supervised learning methods, each with its own advantages and limitations.

\subsubsection{Unsupervised learning methods} \label{methods.unsupervised}

Unsupervised methods, which do not rely on labeled training data, attempt to identify patterns directly from the input data. This approach is particularly valuable when labeled data is scarce or unavailable, as it allows the system to uncover hidden structures within the text. A fundamental concept in unsupervised methods is the grouping of \emph{syntagmas}, or linguistic units, based on shared properties. For example, in the phrase "the black cat," the words collectively form a noun syntagm. Such syntagms can reveal patterns in text that can be leveraged to extract named entities without prior annotation. In unsupervised learning, syntagms can be identified and grouped based on their shared characteristics, such as word order, syntactic roles, allowing the model to discern patterns and structures within the data without prior labeling.

\citet{shinyama2004named} demonstrated this principle by utilizing word distributions to identify named entities, particularly in news articles where named entities frequently co-occur. This method capitalizes on the tendency of named entities to appear together in multiple documents, distinguishing them from common nouns. \citet{Ludovic-Patrice-Michel}, calculated a semantic proximity score by comparing word distributions in documents linked to an entity and the type of entity. \citet{nadeau2006unsupervised} introduced an unsupervised system to build gazetteers and to resolve the ambiguity of named entities. Inspired by previous works such as \citep{etzioni2005unsupervised}, their approach involves amalgamating extracted gazetteers with publicly accessible gazetteers, and the resulting performance has been commendable. However, unsupervised learning methods face several limitations. First, the absence of supervision makes it difficult to assess the accuracy of the extracted named entities, where labeled data could potentially provide a clearer assessment of accuracy. Second, although word distribution methods capture named entities well, they may fail in disambiguating entities with similar surface forms but with different meanings. Finally, the complexity of syntagmatic structures means that unsupervised methods might overlook nuanced semantic differences between entities, leading to less precise groupings or associations.

\subsubsection{Semi-supervised learning methods} \label{methods.semi.supervised}

These methods use labeled and unlabeled data to improve the effectiveness of the model. Unlike traditional supervised approaches that rely exclusively on labeled data, semi-supervised methods leverage the additional information available in unlabeled data to improve the performance of NER systems. Such methods learn from a small labeled dataset using a set of rules designed to identify extraction patterns based on a set of relevant markers. They then attempt to find other samples of named entities that are adjacent to these markers. Subsequently, the learning process is applied to the new samples to discover additional contextual markers. Repeating the process may then lead to the identification of a large number of entities. \citet{collins-singer-1999-unsupervised} demonstrated that a set of seven seed rules, coupled with unlabeled data, can be sufficient to improve model performance in a semi-supervised context.

One approach that is frequently adopted in semi-supervised NER involves co-training \citep{kozareva2005self}, where multiple classifiers are trained on different data views, and these classifier then serve to iteratively label unlabeled instances in order to build a more robust model. An alternative approach involves self-training \citep{gao2021pre}, where the initial model is trained on the labeled data and then used to generate the labels for the unlabeled data. The most confident predictions are added to the labeled data and the process is iterated.

Semi-supervised learning methods offer numerous advantages, such as reducing the dependence on annotated datasets and improving outcomes in low-resource contexts. However, they also present challenges, among which we may mention the risk of propagating errors from the initial labeled data, and the requirement that handling noisy or inaccurate predictions from the unlabeled data is handled meticulously. Despite these challenges, when used judiciously, semi-supervised learning approaches can significantly improve NER performance, and they can have clear benefits in the real world, where labeled data may be scarce or costly to acquire.

\subsubsection{Supervised learning methods} \label{methods.supervised}

These methods rely on patterns derived from labeled data, which require human effort to annotate a set of samples. The labeled samples provide a basis for the model's learning process. The effectiveness of supervised learning methods depends on the quantity and quality of the labeled data used for training.

In the context of NER, the task can be divided into two main subtasks: classification and sequence labeling. During classification, the model learns to identify which words or phrases belong to specific categories of named entities. Sequence labeling, on the other hand, involves assigning a label to each token in a sentence, indicating whether it is part of a named entity and what type it is.

Several prominent models have been developed for NER tasks. These include the Hidden Markov Model (HMM) \citep{baum1970maximization}, which uses probabilistic methods to predict the sequence of labels \citep{zhou2002named, morwal2012named, bikel1999algorithm, zhao2004named}. The Maximum Entropy model (MaxEnt) \citep{berger1996maximum} focuses on finding the most likely label for a given word or phrase based on its context \citep{borthwick1999maximum, curran2003language, chieu2003named, lin2004maximum,borthwick1998nyu}. Support Vector Machines (SVM) are used for classification tasks, separating named entities from non-named entities with a clear margin \citep{cortes1995support, makino2002tuning, ekbal2010named, ju2011named}. Finally, Conditional Random Fields (CRF) \citep{lafferty2001conditional} consider the context of the entire sentence to make predictions about named entities \citep{10.3115/1119176.1119206, settles2004biomedical, nongmeikapam2011crf,shishtla2008experiments}.

\paragraph{\textbf{Hidden Markon Model (HMM)}}

An HMM is a probabilistic model in which the system is assumed to operate as a Markov process. In the context of NER, HMM is used to identify and categorize named entities within a sequence of tokens. In this model, the observed tokens correspond to the observable states, while the various entity labels are regarded as hidden states. The model assumes that the observed tokens depend solely on the current hidden state, which allows the most probable sequence of named entity labels to be inferred based on the observed tokens. Mathematically, an HMM is described by five parameters:
\begin{equation}
    \text{HMM} = \left\{ S, O, \pi, T, E \right\}
\end{equation}
where $S$ represents the number of hidden states (entity labels), $O$ represents the number of observations (tokens), $\pi$ is the initial state probability distribution, $T$ is the transition probability matrix, and $E$ is the emission probability matrix.
The NER problem can be reframed as an HMM problem and expressed as:
\begin{equation}
        P(S|O) = P(N|T).
\end{equation}
This equation posits that given a sequence of tokens $T$, the probability of identifying the sequence of named entities $N$ conditional on $T$ is the same as the probability of identifying the sequence of hidden states $S$ conditional on observations $O$. \citet{bikel1999algorithm} introduced IdentiFinder, an early HMM-based system that effectively learns to identify and categorize names, dates, times, and numerical quantities. The methodology was further developed in subsequent studies, such as \citep{zhou2002named, morwal2012named, zhao2004named}.

\paragraph{\textbf{Maximum Entropy (MaxEnt)}}

MaxEnt models correspond to an advanced statistical method that is often used in NLP tasks, including NER. The core principle of these models is to determine a probability distribution over potential outcomes by maximizing entropy, subject to a predefined set of empirical constraints. The resulting distribution, characterized by having the highest entropy allowable under the observed constraints, is uniquely determined and coincides with the maximum likelihood distribution. It can be expressed as follows:
\begin{equation}
    P(O|H) = \frac{1}{Z(H)} \prod_{j=1}^{k} \alpha_{j}^{f_j(H, O)}
\end{equation}
where $O$ refers to the outcome, $H$ the context, $Z(H)$ is a normalization function, and $\alpha_{j}$ represents the weights corresponding to the features $f_j(H, O)$. The constraints are typically derived from the training data, and the model seeks to assign higher probabilities to the outcomes that are more likely given the observed data. In the context of NER, a MaxEnt model can be trained to predict the label of a named entity for a given token by considering its surrounding context and other relevant features. These features may include information about the current token, its neighboring tokens, and Part-of-Speech (POS) tags, among other elements. During training, the model learns the weights assigned to these features in order to optimize its predictions.

One of the first systems to use the MaxEnt model was presented by \citet{borthwick1998nyu}, giving it the name \textit{Maximum Entropy Named Entity} (\textit{MENE}). A versatile object-based architecture allowed the integration of a wide range of knowledge sources in order to make tagging decisions, and demonstrated of the effectiveness of MaxEnt models in handling NER tasks using diverse contextual information.

\paragraph{\textbf{Conditional Random Fields (CRF)}}

CRFs are probabilistic models widely used for sequence labeling tasks, such as sentence annotation. CRFs take account of the interdependence of neighboring elements, which makes these models exceptionally effective for NER. By capturing sequential dependencies among tokens within a sequence, CRFs adeptly encapsulate the complex contextual relationships characteristic of named entities. Through the integration of both local and global information, CRFs facilitate the prediction of entity labels by considering adjacent token labels, thereby significantly reducing labeling ambiguity. A CRF model is expressed as follows:
\begin{equation}
    P(Y|X) = \frac{1}{Z_0} \exp \Big\{\sum_{t=1}^{T} \sum_{k=1}^{K} \lambda_k f_k (y_{t-1}, y_t, x, t)\Big\}
\end{equation}
where $Z_0$ is the normalization factor for all possible sequences of states (labels), $f_k$ are feature functions, each representing the occurrence of a specific combination of observations and associated labels,
$y_{t-1}$ is the label of the previous word, $y_t$ is the label of the current word, and $x_t$ is the word at position $t$ in the observed sequence.
$\lambda_k$ are the model parameters and can be interpreted as the importance or reliability of the information provided by the binary function.

The NER problem can be formulated as a CRF, where the observations are processed strings, and the labels correspond to the possible named entities. The best sequence of named entities will thus correspond to some sequence of tokens, and finding this best sequence of named entities is equivalent to finding the best sequence of labels, i.e.,  argmax $P(Y = y|X = x)$. \citet{shishtla2008experiments} implemented a system that extracts information from research articles using CRF. They investigated regularization problems using the Gaussian model and focused on the efficient use of feature space with CRF. Settles \citep{settles2004biomedical} presented a framework to recognize biomedical entities using CRF with a variety of features. They demonstrated that a CRF with only simple orthographic features could achieve good performances.

\paragraph{\textbf{Support Vector Machine (SVM)}}

SVMs are a class of machine learning algorithms commonly used for classification tasks. Although SVMs are not as widely used for NER as some other approaches, such as CRF models, they can still be applied effectively with appropriate feature engineering and considerations. \citet{yamada2002japanese} introduced a SVM-based NER system for Japanese, based on Kudo's system \citep{kudo2001chunking}.

In an SVM-based NER system, each word in a sentence is classified sequentially, either from the beginning or the end of the sentence. To handle contextual dependencies, these systems typically incorporate a variety of features that capture the context around each word. These features can include:
\begin{itemize}
    \item \textbf{Word-level features:} The actual word, its part-of-speech (POS) tag, and orthographic features (such as capitalization, presence of digits, etc.) \citep{asahara2003japanese}.
    \item \textbf{Window features:} Information from surrounding words within a fixed-size window (e.g., the previous and next two words) \citep{ju2011named}.
    \item \textbf{Sentence-level features:} Global features that provide information about the sentence structure, such as sentence length or the position of the word in the sentence \citep{takeuchi2002use}.
    \item \textbf{Lexicon features:} Features derived from external knowledge bases or gazetteers that help in identifying named entities based on predefined lists \citep{ekbal2010named}.
\end{itemize}

These features, taken together, allow the SVM to consider the broader context of each word, which is crucial for accurately identifying the named entities. Moreover, by sequentially classifying each word and potentially applying post-processing steps such as Viterbi decoding \citep{viterbi1967error}, SVM-based systems, though not optimized for sequence prediction, can achieve effective NER performance through careful feature engineering and contextual post-processing steps.

Thus, while SVMs might not model sequence dependencies in themselves as well as CRFs do, careful feature engineering and a strategic use of contextual information can nevertheless enable them to perform NER tasks competently.

\paragraph{\textbf{Combined techniques}}

NER systems will often use separate supervised approaches in combination. \citet{srihari2000hybrid} described a hybrid strategy based on MaxEnt, HMM, and custom grammatical rules. Rule-based tagging is used for predictable patterns such as time and monetary expressions, while statistical models such as HMM handle variable entities such as names, locations, and organizations. MaxEnt is used in conjunction with features enriched by external gazetteers to improve tagging accuracy.

\citet{srivastava2011named} presented a combined NER system for Hindi that integrates CRF, MaxEnt, and rule-based methods. This solution addresses language challenges that are peculiar to Hindi, including the absence of capitalization and significant morphological complexity. A voting mechanism merges outputs from CRF and MaxEnt models with custom linguistic rules. \citet{chiong2006named} introduced a sequential hybrid system that uses MaxEnt to initially label named entities within a corpus, providing training data for HMM to finalize the tagging. This method takes advantage of MaxEnt's strength in managing sparse data and the sequential modeling capabilities of HMM. Tests on the British National Corpus demonstrate levels of precision and recall that compare favorably with individual statistical models.

Hybrid methods have proven particularly valuable in low-resource languages, where they can effectively leverage both linguistic rules and machine learning models to improve NER accuracy. For example, a hybrid NER system for Punjabi, proposed by \citet{bajwa2015hybrid}, integrates rule-based methods with HMM. Developed without an existing dataset, the system involved manual tagging to create training and testing data under linguistic supervision. Two versions of NER were introduced: one using only HMM, and another combining HMM with hand-crafted rules. Similarly, to address the lack of existing resources for Arabic, \citet{shaalan2014hybrid} proposed a hybrid NER system combining rule-based methods with machine learning approaches. This system recognizes 11 types of entity, including Person, Location, and Organization, by integrating decision trees \citep{quinlan1986induction}, SVM, and logistic regression classifiers \citep{cox1958regression}.
\subsection{Deep-learning-based methods} \label{methods.deep}

With the rise of deep learning, NER systems have seen significant advances in both accuracy and flexibility. Deep learning methods, which rely on neural networks, have proven particularly effective in automatically learning representations of entities from large datasets. These methods often employ architectures such as Convolutional Neural Networks (CNNs) \citep{lecun1998gradient} and Recurrent Neural Networks (RNNs) \citep{bengio1994learning} to capture both local and sequential patterns within the text.

The introduction of neural probabilistic language models by \citet{bengio2003neural} laid the foundation for deep learning in NLP. By demonstrating the power of distributed word representations, Bengio's work showed that neural networks could capture word similarities in high-dimensional spaces, a capability that would later be crucial for deep learning-based NER systems. Building on this foundation, \citet{collobert2011deep} applied deep learning to NER, proving that CNNs could be used successfully for a wide range of NLP tasks, including NER, semantic role labeling, and chunking.

\begin{figure}[ht!]
  \centering
  \includegraphics[width=2.2in]{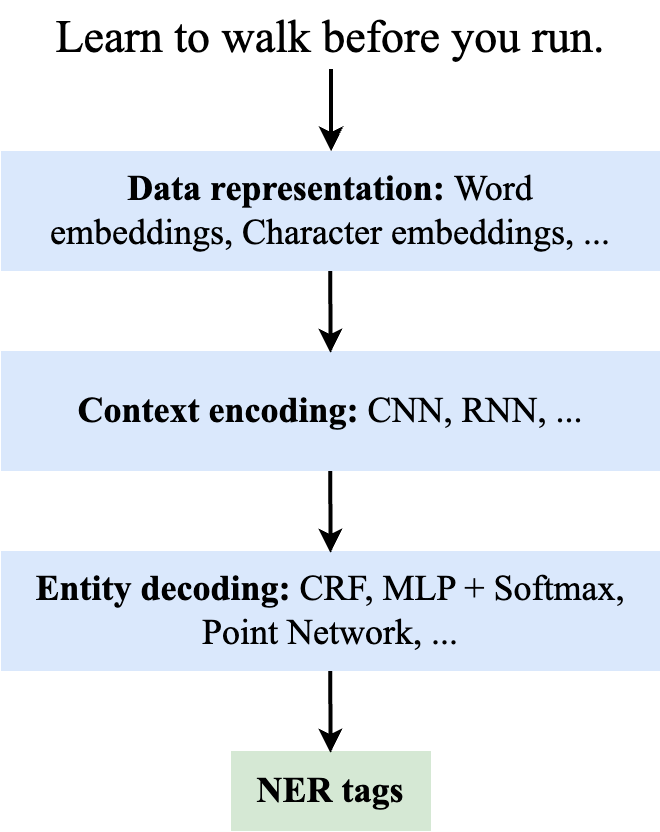}
  \caption{Overview of a deep learning NER model pipeline.}
  \label{deep.learning.arhcitecture}
\end{figure}

Deep learning methods for NER generally follow a three-step process, as shown in Figure~\ref{deep.learning.arhcitecture}: data representation, context encoding, and entity decoding. Each of these steps plays a critical role in ensuring that the model can accurately identify and classify entities within a text.

\subsubsection{Data representation}

Before a deep learning model can process text, the data must be transformed into a format that the model can understand. This requirement can be met through the use of textual embeddings. Numerous techniques for textual embeddings in the context on NER exist, and they can be grouped into two main types: word embeddings and character embeddings.

Word embeddings transform words into dense (or sparse) vectors of real numbers in a high-dimensional space. The resulting vectors represent each word's semantic meaning based on surrounding words in a corpus. Traditional methods include techniques such as One-Hot encoding and TF-IDF \citep{ramos2003using}. Static word embeddings, such as Word2Vec \citep{mikolov2013efficient}, GloVe \citep{pennington2014glove}, and fastText \citep{bojanowski2016enriching}, provide a fixed vector for each word regardless of its context. Contextual embeddings, based on transformers such as GPT \citep{radford2018improving} and BERT \citep{devlin2018bert}, capture the meaning of words based on their specific context within a sentence, allowing for more nuanced and flexible word representations.

Traditional methods for word representation focus on encoding words based on their frequency or presence in a corpus. Although semantic relationships between words are fundamental, these traditional approaches often struggle to capture these relationships, treating each term independently of its context. Two widely used traditional techniques for NER are One-Hot Encoding and TF-IDF. One-Hot encodings represent each word as a sparse binary vector with one component set to 1, indicating the word's presence in the vocabulary. In NER, One-Hot encoding is used to automatically generate training data from sources such as social media \citep{9286444}. TF-IDF is a method that evaluates the importance of a word in a document relative to a collection of documents. It combines two metrics: Term Frequency (TF), which counts how often a word appears in a document, and Inverse Document Frequency (IDF), which measures how rare the word is across the document set. Multiplying TF by IDF highlights important terms that appear frequently in a document but are rare in the general corpus. In NER, TF-IDF can be used to represent features \citep{Abdessalem_Karaa_2011}. The fact that named entities, often rare, receive higher TF-IDF scores as a result of their frequent occurrence within a document but their rare occurrence across the corpus helps them stand out in feature vectors, thus improving entity recognition.

Modern word embeddings like Word2Vec and GloVe produce dense vector representations, where similar terms have close vectors. These embeddings are derived from large corpora in an unsupervised manner. Word2Vec uses Continuous Bag-of-Words (CBOW), which predicts a target word from its context, and Skip-gram, which predicts context words from a target word. Both architectures are trained with noise contrastive estimation (NCE), a type of negative sampling. NCE increases the likelihood of the target word in context and decreases the likelihood of noise words, effectively teaching the model to distinguish between true contexts and artificially generated noise \citep{mikolov2013distributed}. This contrastive approach leads to dense embeddings that are highly effective for capturing semantic similarities, as it forces the model to focus on discriminative features during training, enhancing the quality of the resulting vector representations. GloVe, on the other hand, uses matrix factorization of the word co-occurrence matrix, capturing the corpus's general statistics for efficient, high-performance embeddings \citep{pennington2014glove}. Unlike Word2Vec, which is context-based, GloVe emphasizes global word co-occurrence to produce vectors that represent the global relationships between words.

Numerous studies, including \citep{collobert2011natural, huang2015bidirectional}, have used word embeddings for NER. For example, \citet{ma-hovy-2016-end} evaluated the performance of their NER system using different word embeddings such as Word2Vec and GloVe. The importance of these embeddings to achieve good performance is presented in \citep{lample2016neural}.

Finally, we note that word embeddings can also be combined, as in \citep{dadas2019combining, das2017named}, where a Wikipedia knowledge base is used to annotate named entities. In \citep{dadas2019combining}, the labels are transformed into One-Hot vectors and concatenated with Word2Vec or ELMo \citep{peters2018deep} word embeddings.

\begin{figure}[ht!]
  \centering
  \includegraphics[width=2.7in]{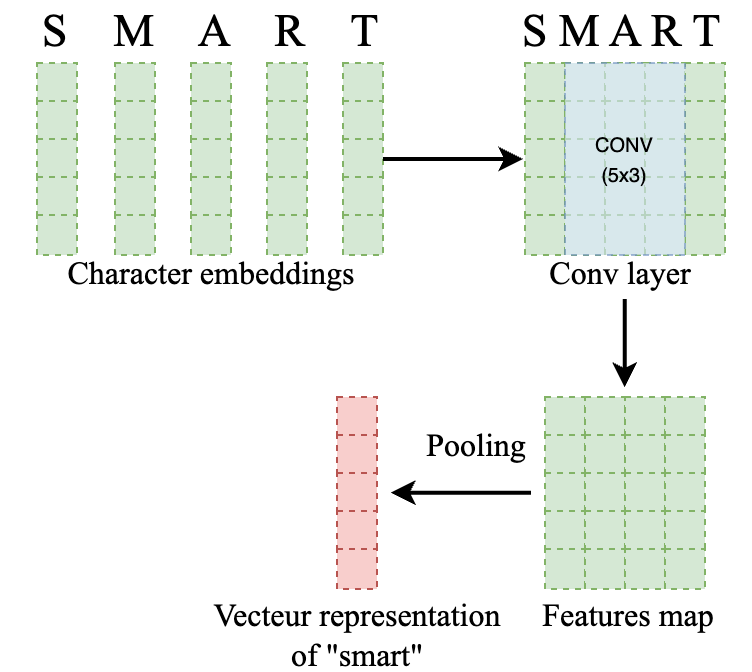}
  \caption{A CNN-based illustration of character embeddings.}
  \label{char.CNN}
\end{figure}

Character embeddings represent words as sequences of character vectors, capturing their internal structure. Unlike traditional embeddings, they vectorize characters and combine them to form word representations, as Figure \ref{char.CNN} illustrates in the case of a CNN architecture.

Spelling variations play a crucial role, as they can reveal the presence of named entities. By representing individual characters, it is possible to capture differences in spelling indicative of a word's syntax and morphology. Character embeddings can also create representations for words not seen during training, known as out-of-vocabulary (OOV) words, by merging various character vectors to form the word.

Character embeddings can be created using several methods. One-hot encoding is a technique that represents characters as binary vectors with a 1 at the index corresponding to the character. Chars2Vec\footnote{https://github.com/IntuitionEngineeringTeam/chars2vec?tab=readme-ov-file}, inspired by Word2Vec, generates embeddings by predicting characters based on their neighboring characters. CNN-based embeddings treat character sequences as images, using CNN filters to extract meaningful features. BiLSTM-based embeddings use bidirectional Long Short-Term Memory \citep{hochreiter1997long, graves2012long} networks to capture contextual information from character sequences.

CNN and RNN, which we describe in the following section, are used in \citep{ma-hovy-2016-end, peters-etal-2017-semi,zhou2023} to calculate feature vectors for each word. One of the conclusions drawn in \citep{lample2016neural} is that recurrent models tend to prioritize later elements, leading to feature vectors that represent suffixes more strongly than prefixes. Consequently, the authors suggest the use of BiLSTM to more effectively capture the prefix information.

\subsubsection{Context encoding}

Once the data is represented as embeddings, the next step is context encoding, which focuses on capturing the relationships and dependencies between words in a sequence. Textual embeddings, based on words, characters, or both, serve as the foundation for various encoding architectures. Among the most widely used are CNNs, which capture local patterns, and RNNs, which handle sequential dependencies. However, newer models like Transformers, which offer a more advanced understanding of contextual relationships, will be discussed separately in Section \ref{methods.Transformers}.

CNN models were initially used for images \citep{oshea2015introduction}, employing filters to detect patterns. They have since been successfully extended to NLP tasks like NER, representing text as word embeddings or characters. This transition from image processing to text is an illustration of CNNs' flexibility in its ability to capture local patterns in different types of data. CNN captures contextual information and local patterns by sliding convolutional layers over the input and using multiple filters to recognize common patterns such as suffixes, prefixes, and word combinations that are indicative of named entities.

A number of researchers have used CNN for NER tasks. \citet{collobert2011natural} introduced a sentence-based network for word tagging, taking into account the whole sentence. Each word is encoded as a vector, and a convolutional layer extracts local features around every word. A global feature vector is created by merging these local features into a fixed dimension, independent of the length of the sentence. This approach of combining local features allows the network to capture both granular and holistic patterns within the sentence, a critical aspect for NER. The global features are then placed in a tag decoder for tag prediction (refer to Section \ref{decoding}). In \citep{gui2019cnn}, the authors presented a CNN-based method for Chinese NER that incorporates lexicons and a rethinking mechanism \citep{li2018learning}. Instead of making a final decision in one pass, the rethinking mechanism includes feedback connections. These feedback loops enhance the model's decision-making by allowing it to reassess its predictions, which helps in refining difficult cases. The connections allow the network to reassess decisions by incorporating high-level feedback into feature extraction. The authors demonstrated that the method can simultaneously model characters and potential words, and the rethinking mechanism can resolve word conflicts by iteratively refining high-level features. In specialized fields such as biomedical NER, CNN-based models have also shown significant promise. For biomedical NER tasks, \citet{zhu2018gram} proposed a deep learning method named GRAM-CNN. The approach leverages local contexts from n-gram character and word embeddings through CNN. Using the local context around each word, GRAM-CNN can autonomously label words without the need for specific knowledge or feature engineering.

In contrast, RNNs have been extensively employed for NER applications. While CNNs excel at capturing local patterns, RNNs are better suited for tasks that require understanding the sequential nature of text. These networks are adept at handling sequential data, making them ideal for tasks in which understanding the context of each word is vital for precise labeling \citep{Sherstinsky_2020}. For NER tasks, the input text is encoded as a series of embeddings, each word being sequentially put into the RNN. The RNN maintains a hidden state that encapsulates information from preceding words. As the RNN processes each word sequentially, the hidden state is continually updated to retain crucial information. However, RNNs are not without their challenges. RNNs face the challenge of the vanishing gradient problem, which hinders their ability to maintain long-term dependencies. RNN variations such as LSTM \citep{Sherstinsky_2020} and Gated Recurrent Unit (GRU) \citep{chung2014empirical} were developed to address the problem. Both LSTM and GRU networks incorporate gating mechanisms that enhance their ability to retain and manage information over extended sequences, thereby improving their effectiveness for NER tasks.

It is for this reason that \citet{huang2015bidirectional} introduced an LSTM model for NER and showed that adding a CRF layer as a tag decoder can improve performance. This combination of LSTM for sequence modeling and CRF for decoding proved to be a highly effective strategy. In other areas, similar approaches were used by \citet{chalapathy2016investigation} for Drug NER and \citet{zhang2018chinese} for Chinese NER. An RNN based on a BiLSTM framework was used in \citep{huang2015bidirectional}; the same approach has since been adopted by other authors \citep{ma-hovy-2016-end, lample2016neural, hang2020}. This widespread adoption underscores the robustness and flexibility of RNN-based models for a variety of NER applications.

\subsubsection{Entity decoding} \label{decoding}

The final phase of the deep learning pipeline is entity decoding, where the model assigns entity labels to each word in the input sequence. A variety of architectures are commonly used for this task, including CRF, Multi-Layer Perceptron (MLP) and Pointer Networks \citep{vinyals2015pointer}. Each architecture has its strengths, depending on the complexity of the NER task.

CRF, for example, is a probabilistic model that assigns labels by considering the entire sequence of tokens rather than making independent predictions for each one. It excels in modeling dependencies between adjacent labels, which is crucial for ensuring coherent tag sequences. By learning to assign higher probabilities to valid label sequences (such as recognizing that a "Person" label is likely to be followed by a "Location"), CRF provides a robust method for NER. Studies by \citet{lample2016neural} and \citet{ma-hovy-2016-end} show that incorporating a CRF layer into deep learning models, such as BiLSTM \citep{luo2018attention, lin2019reliability} or CNN \citep{knobelreiter2017end, feng2020study}, enhances performance. This combination leverages CRF’s ability to ensure consistent sequence labeling while benefiting from the feature extraction power of neural networks. Many advanced NER models integrate a CRF layer with either BiLSTM or CNN for feature extraction. The BiLSTM captures the sequential context by processing the input in both forward and backward directions, ensuring that the model considers the entire sentence before making predictions. CNN, on the other hand, focuses on local patterns such as prefixes, suffixes, and word combinations, which are key in identifying entities. The CRF layer is then used to refine the final output, ensuring consistency across the sequence. This combination allows the model to capture both local features and long-range dependencies, making it more accurate and reliable.

In contrast, MLP, a simpler architecture, assigns entity labels independently to each word. The MLP transforms the sequence labeling task into a multiclass classification problem by using a softmax layer to predict the tag of each token separately. Although this approach is easier to implement and can work well for basic NER tasks, it lacks the ability to capture relationships between adjacent tokens. This limitation means that MLP models, although effective for straightforward cases, may underperform in more complex contexts where understanding token dependencies is critical, as noted by \citet{gallo2008named,lin2019reliability}.

For more dynamic tasks, Pointer Networks represent a different approach. They are designed to handle variable-length output sequences and use attention mechanisms to directly point to elements in the input sequence rather than relying on a fixed set of output labels. This approach provides greater flexibility, particularly when dealing with sequences that involve unknown or variable output lengths, which is often the case in NER tasks. By computing soft alignment scores between input and output elements, Pointer Networks can dynamically adjust to different sequence structures, making them highly effective in tasks that require adaptable and context-specific output \citep{zhai2017neural,li2019adversarial,skylaki2020named}.

Finally, recent innovations in NER, such as the approach introduced by \citet{fei2021rethinking}, focus on token-to-token interactions rather than traditional sequential tagging. This method shifts the emphasis from token-level prediction to understanding the relationships between pairs of tokens, which allows for a deeper understanding of entity relations within a sentence. Using a multiview contrastive learning framework that aligns both semantic and relational spaces across languages, this model improves multilingual NER tasks. This evolution in NER methodologies highlights the importance of capturing more complex relationships between tokens, offering improved performance across different languages and contexts.

\subsection{Transformer-based language methods} \label{methods.Transformers}

Transformer-based models have significantly reshaped NLP tasks, and in particular NER, by allowing the modeling of complex contextual relationships within text. As described in \citet{vaswani2017attention}, the Transformer model utilizes self-attention mechanisms to effectively capture and integrate contextual information. Transformer-based encoder models, such as BERT, have significantly influenced NER. These architectures undergo extensive pre-training on large corpora, followed by fine-tuning on domain-specific NER datasets. BERT, in particular, has demonstrated high efficacy across various NER tasks due to its ability to produce richly contextualized embeddings. Empirical studies have shown that using BERT as a classifier consistently outperforms traditional BiLSTM-CRF architectures \citep{riedl2018named, schweter-baiter-2019-towards,oralbekova2024comparative}.

Several BERT derivatives, notably DistilBERT \citep{sanh2020distilbert} and Robustly Optimized BERT Pre-training Approach (RoBERTa) \citep{liu2019roberta}, have demonstrated strong performance in NER tasks \citep{abadeer2020assessment,mehta2021named,su2022roberta,hofer2023minanto}. DistilBERT, a smaller and faster variant of BERT, retains a substantial portion of BERT's language understanding capabilities. On the other hand, RoBERTa, trained on a larger dataset and benefiting from improved training techniques, generates deeper contextual embeddings and achieves superior performance across various NLP tasks, including NER. Decoding-enhanced BERT with Disentangled Attention (DeBERTa) \citep{he2020deberta} offers an alternative model that has demonstrated substantial improvements in NER and various other NLP tasks. DeBERTa integrates disentangled attention mechanisms and an enhanced mask decoder, allowing for more efficient capture of word dependencies and increased model robustness. This sophisticated pre-training has enabled DeBERTa to outperform both BERT and RoBERTa on several evaluation benchmarks. PLTR \citep{wang2023generalizing} employs pre-trained models such as RoBERTa to generate contextual embeddings for NER tasks, specifically avoiding recent LLMs like GPT due to their high computational costs and relatively poor performance in NER. By incorporating prompts and type-related features, PLTR improves the model's ability to generalize across domains. Another notable approach is TDMS-IE, proposed by \citet{hou2019identification}. This method automates the extraction of information such as tasks, datasets, and evaluation metrics from scientific papers using BERT. By combining ontology-based and data-driven techniques, TDMS-IE excels at identifying emerging methodologies, which makes it particularly useful for constructing leaderboards in NLP research.

Generalist Model for NER using Bidirectional Transformer (GLiNER) \citep{zaratiana2023gliner} constitutes a novel paradigm that integrates global contextual information into NER. The model employs a global attention mechanism that enables it to adeptly utilize long-distance entity relationships. This capability is especially advantageous in dealing with complex NER tasks that require contextual understanding that span multiple sentences. Additionally, Transformer-based models have been adapted for specific languages and domains. For example, \citet{choudhry2022Transformerbased} proposed an approach for French using adversarial adaptation to overcome the lack of labeled NER datasets. By training the models on labeled source datasets and utilizing larger corpora from other domains, they succeeded in improving feature learning.

Although Transformer-based models such as BERT, RoBERTa, DeBERTa, and GLiNER have shown excellent performances in NER, they are computationally intensive. The motivation behind attempts to create more efficient models such as DistilBERT is obtaining a better balance between performance and resource requirements. Recent advances also include the development of multilingual models such as mBERT \citep{devlin2018bert}, which can handle multiple languages and be fine-tuned for specific languages using bilingual gazetteers or additional datasets, enhancing performance in low-resource settings. Additionally, models such as Language Understanding with Knowledge-based Embeddings (LUKE) incorporate entity information into word embeddings, improving entity recognition by using innovative masking and self-attention techniques \citep{yamada2020luke}.

These models can also be combined with other architectures to further enhance NER performance. For example, combining BERT with LSTM networks helps capture both long-term dependencies and contextual information \citep{souza2019portuguese, wan2020research, he2021named, chen2021joint}. BERT provides robust contextual embeddings, which LSTM processes to model sequential dependencies more effectively. Integrating BERT with BiLSTM improves the model's ability to capture dependencies in both forward and backward directions, improving the accuracy of NER tasks by leveraging comprehensive context from both directions \citep{dai2019named, chang2021chinese, xu2021named, lee2022ncuee, shi2024bert}. Using CNN with BERT can capture local patterns in the text, as the convolutional layers detect local features in the BERT embeddings, which can be particularly useful for identifying entities in specific contexts or within fixed-size windows of text.

Combining BiLSTM and CNN architectures within the same NER framework can also significantly enhance performance. In the hybrid approach, BERT embeddings are first fed into BiLSTM layers to capture long-term dependencies and bidirectional context. The output of the BiLSTM layers is then processed through CNN layers to identify local patterns and features. The combination leverages the strengths of the two architectures, that is to say BiLSTM's ability to model sequential and contextual information and CNN's efficiency in detecting local patterns. The dual architecture improves the overall performance of the NER by capturing a wide range of dependencies and contextual cues from the text \citep{wu2022one,chen2022named}.

It is, moreover, possible to effectively integrate other RNN architectures such as GRU into NER frameworks. GRUs offer a simpler and more computationally efficient alternative to LSTM while still maintaining the ability to capture long-term dependencies \citep{alsaaran2021arabic}. Integrating BERT embeddings with GRU networks can yield similar benefits as with LSTM networks. Combining GRUs with CNN layers can further enhance the ability to detect local patterns and features in the text, resulting in improved NER performance.

\subsection{Large language model-based methods}
\subsubsection{Principles and applications of LLMs in NER}
LLMs constitute an advanced category of deep learning architectures that have the capacity to perform various tasks, including, but not limited to, translation, summarization, classification, and content generation. These models are characterized by their substantial numbers of parameters, which will often extend into the tens or hundreds of billions. They are trained on large datasets, such as GPT \citep{brown2020language}, BloomZ \citep{muennighoff2023crosslingual}, and LlaMA \citep{touvron2023llama}.

LLMs are based on the Transformer decoder architecture, in which a multitude of attention mechanisms are orchestrated in layers to form an intricate neural network. The structural designs and pre-training paradigms implemented in current LLMs exhibit strong parallels with those used in smaller-scale language models. The primary distinction is the markedly increased size of both the model parameters and the training corpus. Some LLMs, such as T5 \citep{2020t5}, operate as hybrids, incorporating the encoder and decoder modules of the Transformer to increase comprehension and generative functionalities.

\begin{table}
\caption{Summary of studies on LLMs in NER}
\centering
\begin{tabular}{p{3.5cm}|p{4.5cm}|p{7cm}}
\hline
\textbf{Study} & \textbf{Approach} & \textbf{Outcome} \\
\hline
GPT-NER \citep{wang2023gptner} & Transforms sequence labeling to text generation & Comparable to fully supervised baselines, better in low-resource and few-shot setups \\
\hline
PromptNER \citep{ashok2023promptner} & Uses entity type definitions for few-shot learning & State-of-the-art performance on few-shot NER, significant improvements on various datasets \\
\hline
ChatGPT Evaluation \citep{laskar2023systematic}	 & Evaluates ChatGPT on various NER tasks & Impressive in several tasks, but far from solving many challenging tasks \\
\hline
Injecting comparison skills in TOD Systems \citep{kim2023injecting}	 & Compares properties of multiple entities & Effectively addresses ambiguity handling in database search results \\
\hline
Zero-Shot on historical texts with T0 \citep{de2022entities} & Explores zero-shot abilities for NER & Shows potential for historical languages lacking labeled datasets, error-prone in naive approach \\
\hline
Resolving ECCNPs \citep{kammer2023resolving} & Proposes a generative encoder-decoder Transformer & Outperforms rule-based baseline \\
\hline
Large code generation models \citep{li2023codeie} & Uses generative LLMs of code for Information Extraction tasks & Consistently outperforms fine-tuning moderate-size models and prompting NL-LLMs in few-shot settings \\
\hline
UniversalNER \citep{zhou2023universalner} & Targeted distillation from LLMs & Broad coverage of entity types, suitable for clinical applications \\
\hline
Self-Improving Zero-Shot NER \citep{xie2023self} & Unlabeled corpus for self-improvement & Enhanced zero-shot capabilities through self-annotated pseudo-demonstrations \\
\hline
GL-NER \citep{zhu2024gl} & Generation-aware LLM with label-injected instructions & Improves few-shot learning performance with novel prompt template and masking-based loss \\
\hline
E2DA \citep{zhang2024exogenous} & Combines exogenous and endogenous data augmentation & Significantly improves performance in low-resource contexts \\
\hline
GPT-4 and Claude v2 \citep{chebbi2024enhancing} & LLMs applied to dynamic entity extraction & Adapts to new entity types in dynamic environments \\
\hline
CALM \citep{luiggi2024} & Generates additional context for entities offline & Creates relevant context to improve low-resource settings \\
\hline
\end{tabular}
\end{table}

LLMs are lauded for their exceptional performance in various NLP tasks, including text classification \citep{hegselmann2023tabllm}, question answering \citep{robinson2022leveraging}, text generation \citep{muennighoff2023crosslingual}, and machine translation \citep{hendy2023good}. However, their application to NER, a sequence labeling task, has revealed some limitations, as LLMs were originally designed for text generation.

To address the gap between LLMs and NER, various innovative methods have been developed. One notable method is GPT-NER \citep{wang2023gptner}, which converts sequence labeling into a text generation task. For example, rather than directly identifying entities, the task of marking a location entity such as "Paris" is reformulated as generating a modified sentence: "@@Paris\#\# is a city," where the special tokens "@@" and "\#\#" indicate entity boundaries. This transformation allows LLMs to perform sequence labeling in a more natural text generation format, yielding promising results, especially in few-shot and low-resource settings, where training data are limited. Another significant development in NER involves using few-shot learning approaches, which allow models to learn from a minimal number of examples. PromptNER \citep{ashok2023promptner} shows how entity type definitions in prompts enable LLMs to list entities with explanations. This method has shown state-of-the-art performance in few-shot NER, demonstrating its ability to generalize across domains with minimal training data. Moreover, research including \citep{hu2023zero, laskar2023systematic} has investigated the potential of ChatGPT in zero-shot or few-shot clinical NER scenarios. Although initially developed for general text generation, ChatGPT has demonstrated performance on par with specialized models such as BioClinicalBERT \citep{alsentzer2019publicly}, although it faces certain challenges in more complex tasks.

To further advance few-shot learning, GL-NER \citep{zhu2024gl} was introduced as a generation-aware LLM specifically designed for few-shot NER. GL-NER employs a novel prompt template that incorporates label-injected instructions, enabling it to either generate entity names or to signal "does not exist" when no entity is present. In addition, it uses a masking-based loss optimization strategy that significantly improves few-shot learning performance over traditional prompt-based methods. This approach helps to tackle the inherent challenges of few-shot NER, where labeled examples are scarce.

Hybrid approaches have also been explored to combine the strengths of different LLMs. For example, BERT and GPT-2 have been used together in order to disambiguate named entities in dialogue systems \citep{kim2023injecting}. In this setup, GPT-2 acts as a generator during the training phase, while BERT performs the evaluation during inference. This combination allows the model to effectively address ambiguity and entity comparison in real-time dialogue, showing the potential of integrating different LLM architectures to tackle specific NER tasks.

In more specialized contexts, such as historical and multilingual NER, an evaluation has been done of the T0 multitask model \citep{de2022entities}. This study highlights the unique challenges presented by historical texts, including language variations and inconsistent spellings. Although the model showed potential, particularly in its ability to identify languages and publication dates, it struggled with zero-shot NER in these specialized domains, where labeled data are often scarce. In medical domains, a successful handling of complex language constructs is crucial. \citep{kammer2023resolving} proposed a generative Transformer model to address the challenge of elliptical compound nominal phrases (ECCNPs) in German medical texts. This method exhibited a significant improvement over rule-based systems, demonstrating the power of LLMs to handle specialized language constructions and to increase accuracy in medical NER tasks.

In the realm of information extraction, \citep{li2023codeie} explored the use of code-based LLMs (Code-LLMs) for tasks traditionally tackled by natural language LLMs. By reframing information extraction tasks as code generation problems, Code-LLMs like Codex \citep{chen2021evaluating} have shown better performance than traditional NL-LLMs in few-shot setups. Code-based LLMs show the potential of using LLMs trained in different modalities (such as code) to improve performance in specific NLP applications such as NER.

Targeted distillation techniques have also gained attention as a way to improve NER in specific domains. UniversalNER \citep{zhou2023universalner} distills the knowledge from LLMs to handle open-domain NER tasks, sampling inputs from large and diverse corpora and using ChatGPT to generate a wide variety of entity types. This makes UniversalNER suitable for applications such as clinical NER, where entity diversity is crucial. Similarly, Self-Improving Zero-Shot NER \citep{xie2023self} introduces a framework for improving performance by leveraging an unlabeled corpus. Through self-annotated pseudo-demonstrations, the model continuously improves its zero-shot capabilities.

Among efforts seeking to improve NER through contextual information, CALM \citep{luiggi2024} involves generating additional context for entities offline. This approach leverages LLMs to create relevant context, especially useful in low-resource settings where the availability of annotated data is limited. \citep{chebbi2024enhancing} applied similar LLM-based techniques in the agricultural sector, using models such as GPT-4 and Claude v2 to monitor agricultural commodities. These models extracted and classified key entities from unstructured sources such as market reports and trade documents. Using prompt-based few-shot learning, LLMs were able to adapt to new entity types in dynamic environments without extensive domain-specific retraining.

Finally, advanced data augmentation techniques have been explored to further improve NER performance. \citep{zhang2024exogenous} introduced E2DA, which combines exogenous and endogenous data augmentation. Exogenous augmentation uses LLMs to generate additional data based on specific instructions, increasing the diversity of the data set, while endogenous augmentation exploits semantic relationships within the data to maximize the use of meaningful features. This method significantly improves NER performance in low-resource contexts.

\subsubsection{Pros and Cons of LLMs for NER}
Recent advances in LLM-based NER approaches underscore several advantages. First, versatility and adaptability stand out as key strengths. LLMs' ability to perform few-shot and zero-shot learning is particularly advantageous in contexts where labeled data are scarce or unavailable. This capability allows these models to generalize with minimal data input, as demonstrated by few-shot models such as PromptNER and UniversalNER, which have shown good performance. This adaptability makes LLMs highly suitable for domains where annotated data are often costly or difficult to obtain.

Another important advantage is contextual understanding. The transformer-based architecture of LLMs enables these models to capture complex contextual relationships, an essential feature for NER tasks in specialized domains, such as medical or historical texts. For example, models such as Resolving ECCNPs \citep{kammer2023resolving}, customized for German medical texts, effectively handle unique language constructs and domain-specific terminology. This ability to understand nuanced context improves LLM performance in scenarios where accurate interpretation of specialized language is crucial.

Finally, LLMs demonstrate significant potential in multi-domain applications. Through prompt engineering and self-improvement techniques, these models have shown promising results across a range of applications, from clinical NER to information extraction in sectors like agriculture. This adaptability is particularly beneficial in fields that require dynamic knowledge, where entity types and contexts frequently change. LLMs such as those used in agricultural commodity monitoring demonstrate the practical value of adapting to new entity types without extensive retraining, which adds to their appeal in diverse applications.

However, these models are not without their challenges. A primary drawback is that LLMs are resource-intensive. The high computational and memory demands for training and deploying these models make them costly and often inaccessible for smaller-scale applications or institutions with limited resources. The large-scale infrastructure required to run models with billions of parameters may limit LLM adoption in resource-constrained environments, despite their advantages in performance. Additionally, LLMs exhibit prompt sensitivity. Many LLM-based NER approaches, such as PromptNER and UniversalNER, rely heavily on prompt engineering. Crafting effective prompts requires substantial fine-tuning and domain expertise, as small changes in prompt design can lead to significant variations in model performance. This dependence on prompt quality introduces a degree of unpredictability and limits the scalability of LLMs, particularly in settings where consistent outputs across tasks are crucial. Lastly, LLMs face limitations in complex or ambiguous contexts. Although generally effective in standard NER tasks, models like ChatGPT struggle with more intricate scenarios, especially when dealing with nuanced or domain-specific entity types. These challenges are pronounced in tasks that require a deep understanding of specific domain knowledge, where generic LLMs may not capture subtle distinctions without extensive domain-specific training. This limitation suggests a need for specialized adaptations or hybrid models to effectively address complex or ambiguous NER tasks.

\subsection{Reinforcement learning}

RL has emerged as a promising approach for increasing the performance and adaptability of NER systems. Agents are trained to make sequences of decisions through the rewarding of desirable actions and the penalizing of undesirable ones. In the context of NER, RL can optimize the identification and classification of named entities by learning from interactions with data and progressively improving the model's performance through rewards and penalties.

Combining RL with entity triggers, \citet{yang2023gaussian} proposed a Gaussian Prior Reinforcement Learning framework (GPRL) to learn the order of recognition of the entity and use the boundary positions of nested entities. GPRL converts nested NER into an entity triplet sequence generation task using BART \citep{shen2021locate} with a pointer mechanism \citep{strakova2019neural}. To improve nested entity recognition, a Gaussian prior adjustment is applied to the probability of the boundary of the entity predicted by the pointer network. The recognition order is modeled as an RL process, optimizing the network by maximizing triplet generation rewards.

Another significant approach is to integrate distant supervision and RL. Distant supervision addresses the scarcity of labeled data by using external resources to automatically generate annotated data sets, but it is a process that will often introduce noise \citep{10.1007/978-3-030-60450-9_27}. RL helps mitigate the noise by employing confidence calibration strategies to refine the model's predictions, thereby improving the overall performance of NER systems.

It is also possible to combine RL with adversarial training. Adversarial training consists in training models to distinguish between true and false positives in a competitive setting, which can improve resilience to noisy and incomplete annotations. In the biomedical domain, RL improves the recognition of complex medical entities, handling a wide range of biomedical terms and variations, thus improving the extraction of meaningful entities from medical texts \citep{10.1007/978-3-030-82136-4_16}.

Studies such as \citep{yang2018distantly} have shown that RL can effectively handle incomplete and noisy annotations. For example, by integrating partial annotation learning, RL reduces the impact of unknown labels. An RL-based instance selector can filter out noisy annotations, further refining the model's accuracy.

The SKD-NER model is an example of advanced techniques in continual learning for NER \citep{chen2023skd}. This model addresses the problem of catastrophic forgetting, where a model trained to identify new entities loses its ability to recognize previously learned ones. SKD-NER uses knowledge distillation to maintain memory and applies RL strategies during this process. It fine-tunes soft labeling and distillation losses produced by the teacher model, effectively mitigating catastrophic forgetting during continual learning.
\subsection{Graph-based methods} \label{methods.graph}
\begin{figure*}[ht]
  \centering
  \includegraphics[width=4.9in]{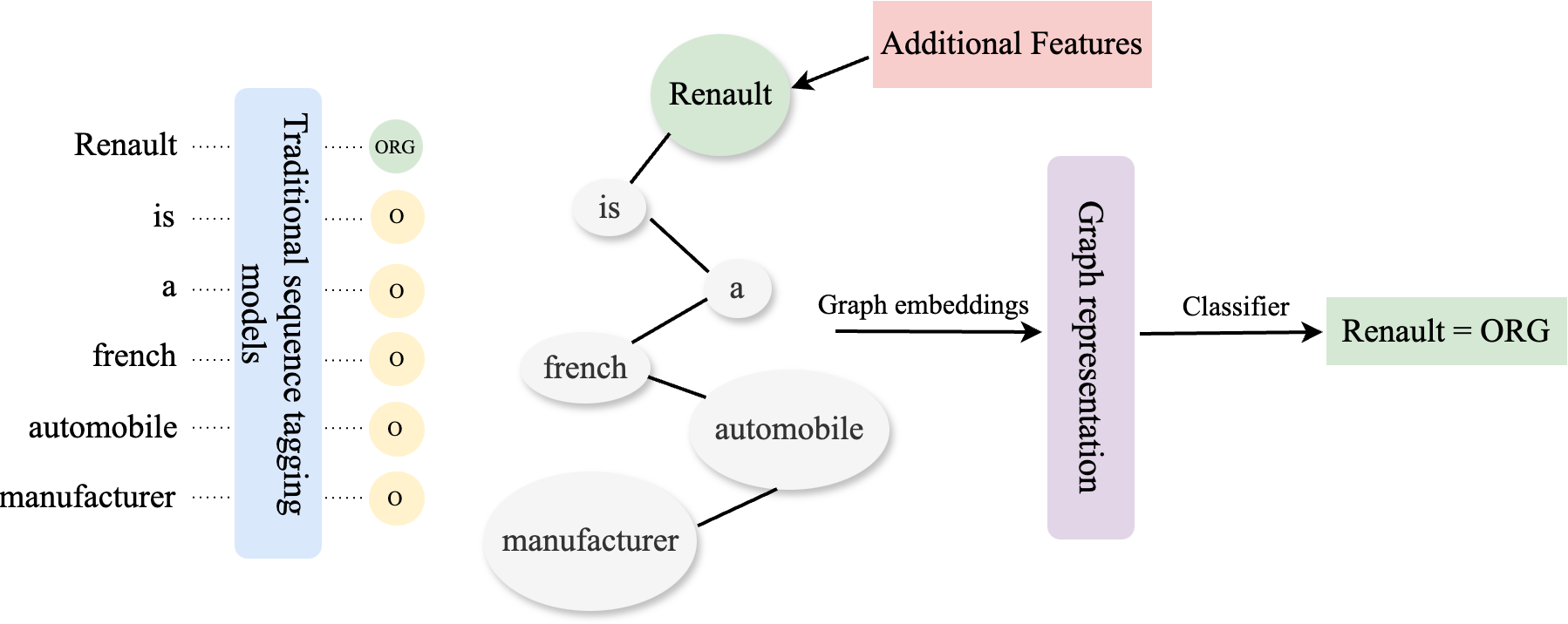}
  \caption{Left side: Conventional model for sequence tagging. Right side: Every word within a sentence transforms into a node of a graph connected to surrounding words and additional features like grammatical characteristics. This graph is subsequently encoded and fed into a classifier to predict entity tags.}
  \label{fig.comp.graph}
\end{figure*}
The use of Graph Convolutional Networks (GCNs) in NLP allows sequences to be represented as graph structures. In these methods, each token in a sentence is treated as a node, and edges represent relationships between them (see Figure \ref{fig.comp.graph}). This approach was first used in NLP by \citet{marcheggiani2017encoding}, leveraging GCNs as proposed by \citet{kipf2016semi}. \citet{cetoli2017graph} demonstrated that applying GCNs to NER tasks was able to improve performance significantly, particularly on the OntoNotes 5.0 dataset \citep{weischedel2013ontonotes}. This representation transforms sequential data into richer structures, capturing contextual dependencies effectively.

Graph embeddings have been widely adopted to strengthen token representation in NER. \citet{liu2019graph} proposed a GCN-based architecture that generates embeddings from graph structures. These graph embeddings were then combined with token embeddings and processed through a BiLSTM-CRF network to predict NER tags. Similarly, \citet{harrando2021named} viewed the NER task as a graph classification problem, where each token in a sequence serves as a node, and features such as morphological shape or POS-tag are incorporated to enrich the node representation. These approaches exhibited better token representations and led to better entity recognition outcomes.

Further advances, particularly for Chinese NER, have led to the development of specialized graph-based networks, such as the Polymorphic Graph Attention Network (PGAT) by \citet{wang2022polymorphic}. PGAT improves character representation by dynamically integrating lexicon information. Another approach, the Lexicon-Based Graph Neural Network (LGN) by \citet{gui2019lexicon}, constructs connections among characters, leveraging both local compositional structures and global sentence semantics. This approach demonstrated significant performance improvements across various Chinese NER datasets. Furthermore, the Multi-Graph Collaborative Network (MGCN) of \citet{10.1007/978-3-031-17120-8_48} addresses challenges like boundary confusion and irrelevant lexical words, outperforming existing models by constructing multiple relationships between lexical words and characters in order to better capture entity boundaries.

Multimodal NER presents additional challenges, such as dealing with both textual and visual data. \citet{zhao2022learning} introduced the Relation-enhanced GCN (R-GCN) for Multimodal NER (MNER), using inter-modal and intra-modal relation graphs to effectively integrate text and image information. This method allows for better identification of entities in cases where information is spread across multiple modalities. Another challenge is embedding newly emerging entities in existing knowledge graphs. The VN Network introduced by \citet{he2020vn} addresses this problem by generating "virtual neighbors" for unseen entities. This approach involves creating connections based on logical rules and utilizing Graph Neural Networks (GNNs) to aggregate information from these synthetic neighbors, thus improving the representation of new entities.


\subsection{Annotation schemes}

NER relies on precise annotation schemes to effectively segment and label entities within a text. These schemes serve as the foundation for determining what constitutes an entity and where it begins and ends within a sentence. Choosing the appropriate annotation scheme is crucial for ensuring that the model can accurately perform the sequence-labeling task. Different schemes are designed to handle the complexity of identifying multi-token entities and distinguishing between entity and non-entity tokens.

In most annotation schemes, the first token of a named entity is tagged as B (Begin), marking the start of the entity. If the entity spans multiple tokens, the intermediate tokens are labeled as I (Inside), while the last token can be tagged as either E (End) or I, depending on the specific schema. Tokens that do not belong to any named entity are assigned the tag O (Outside). Annotation schemes vary in their use of combinations of tags in order to handle different contexts. Commonly used schemes include BIO (Begin, Inside, Outside), IO (Inside, Outside), IOE (Inside, Outside, End), IOBES (Inside, Outside, Begin, End, Single), IE (Inside, End), and BIES (Begin, Inside, End, Single). In some implementations, the End tag is denoted as L (Last) rather than E. Each of these schemes offers different advantages depending on the nature of the entities and the complexity of the text being analyzed.

\begin{itemize}
    \item The IO scheme is the simplest method in which each token receives either an I or an O tag. The I tag represents named entities, while the O tag represents other words. One limitation of this schema is its inability to differentiate between consecutive entity names of the same type.
    \item The BIO schema, widely used and adopted in the CoNLL Conference, assigns one of three tags to each token: B for the start of a named entity, I for inside tags within the entity, and O for outside tags indicating non-entity words.
    \item The IOE schema is similar to BIO, but instead of marking the start of a named entity (B), it denotes the end of the entity (E).
    \item IOBES serves as an extension of the IOB scheme, offering more information regarding the boundaries of named entities. It uses five tags: B for the beginning of an entity, I for inside tags within the entity, E for the end of an entity, S for single-token entities, and O for non-entity words outside named entities.
    \item The IE scheme functions similarly to IOE, with the difference that it labels the end of non-entity words as E-O and the rest as I-O.
    \item The BIES scheme is an extension of IOBES. It utilizes tags such as B-O for the beginning of non-entity words, I-O for inside tags within non-entity words, E-O for the end of non-entity words, and S-O for single non-entity tokens located between two entities.
\end{itemize}

\begin{table}[!ht]
\centering
\small
\caption{Comparison of different annotation schemes on a sample sentence, where "PER" denotes a person and "ORG" represents an organization.}
\begin{tabular}{ccccccc}
\hline
{\bf Words} & {\bf IO} & {\bf BIO} & {\bf IOE} & {\bf IE} & {\bf IOBES} & {\bf BIES} \\
\hline
Emma & I-PER & B-PER & I-PER & I-PER & B-PER & B-PER \\
\hline
Charlotte & I-PER & I-PER & I-PER & I-PER & I-PER & I-PER \\
\hline
Duerre & I-PER & I-PER & I-PER & I-PER & I-PER & I-PER \\
\hline
Watson & I-PER & I-PER & E-PER & E-PER & E-PER & E-PER \\
\hline
was & O & O & O & I-O & B-O & B-O \\
\hline
born & O & O & O & I-O & I-O & I-O \\
\hline
in & O & O & O & E-O & E-O & E-O \\
\hline
Paris & I-ORG & B-ORG & I-ORG & I-ORG & S-ORG & S-ORG \\
\hline
. & O & O & O & I-O & B-O & S-O \\
\hline
\end{tabular}
\label{table:sch}
\end{table}

For a comparison of these annotation schemes, we refer the reader to Table~\ref{table:sch}, which illustrates the application of the different schemes to an example sentence.

It is worth noting that the choice of tag scheme can also affect NER performance. For example, \citet{alshammari2021impact} found that the IO scheme outperforms other schemes for articles written in Arabic. Similarly, \citet{chen2021novel} demonstrated that the IO scheme is more suitable for steel e-commerce data compared to the BIO and BIEO schemes.

\section{Low-resource NER} \label{low.resource}

Although neural networks and Transformer-based models have shown remarkable success in NER, their effectiveness is often tied to the availability of large annotated datasets. However, in many real-world scenarios, especially for low-resource languages or specialized domains, such datasets are either limited or nonexistent. This presents a significant challenge for NER systems, as lack of sufficient data can hinder model training and performance.

To address these challenges, several strategies have been proposed to overcome data scarcity issues in low-resource NER settings. Using these techniques, NER models can be adapted to perform effectively even when faced with limited labeled data, ensuring that entity recognition can be applied in a wider range of contexts and languages.

\subsection{Transfer learning} \label{low.resource.transfer} Transfer learning applies knowledge from one task in order to improve performance on a related task. It is a technique that has proven useful in areas such as image classification \citep{shaha2018transfer}, speech recognition \citep{wang2015transfer}, and time series classification \citep{fawaz2018transfer}. In NER, transfer learning is implemented by first pre-training a model on a large corpus of generic text data, followed by fine-tuning on a smaller dataset specifically designed for the target NER application. This approach utilizes the broad knowledge gained from the generic dataset so as to obtain better performance on the specific task at hand. One example of transfer learning is BERTweet, which is a BERT-based model pre-trained on Twitter data that achieves good performances on Twitter text classification, as well as on POS-tagging and named-entity recognition \citep{nguyen2020bertweet}.

Numerous studies have examined the application of transfer learning to NER. \citet{lee2018transfer} looked at transfer learning through RNN in the context of anonymization of health data. Studies such as \citet{francis2019transfer} have explored the implementation of Transformers for NER, demonstrating that transfer learning significantly improves performance. More recently, \citet{fabregat2023negation} proposed a range of architectures based on BiLSTM and CRF to detect biomedical named entities using negation-based transfer learning techniques. Specifically, this approach incorporates negation detection as a pre-training step, where weights relating to negation triggers and scopes are transferred, resulting in improvements in both NER and Relation Extraction (RE) tasks.

\subsection{Data augmentation} \label{low.resource.data.augmentation}

Data augmentation is a technique that is used to artificially increase the size of a training dataset by creating modified versions of existing data. This can involve applying small transformations to the original data, such as synonym replacement, random deletion, insertion, swap, back-translation, or lexical substitution \citep{dai2020analysis, sawai2021sentence, duong2021review}. Generative models can also be used to synthesize entirely new examples, further enriching the training set \citep{sharma2022systematic, 9207241}.

The application of data augmentation techniques to NLP has been explored in various areas, including text classification \citep{dai2020analysis, karimi2021aeda}, machine translation \citep{sawai2021sentence}, and sentiment analysis \citep{duong2021review}. However, unlike other NLP tasks, NER involves making predictions about words rather than sentences. This brings with it an additional challenge, since applying transformations to words can alter their labels, and this complicates the use of data augmentation for NER. Despite this difficulty, adaptations of data augmentation techniques for NER have been done. For example, \citet{dai2020analysis} applied simple strategies like word replacement and named entity replacement to improve model performance, particularly for datasets containing very few examples. In a similar study, \citet{chen2020local} applied Local Additivity based Data Augmentation (LADA), which generates new samples by combining similar ones. LADA has two variants: Intra-LADA and Inter-LADA. Intra-LADA creates new sentences by exchanging words within a single sentence and interpolating between these new sentences, whereas Inter-LADA combines different sentences to construct new data. More complex strategies, such as paraphrasing, have also been employed to generate new data \citep{sharma2022systematic}.

Data augmentation is a vital technique in training NER models, especially when dealing with limited datasets. In the future, data augmentation in NER might potentially be made much easier and more effective through the use of LLM. LLMs can generate realistic and diverse text data, which can be used to augment existing datasets for NER training. This approach, as exemplified by techniques such as GPT3Mix \citep{yoo2021gpt3mix}, allows for the creation of more accurate NER models by enriching training data with a wide range of linguistic variations and contextual scenarios.

\subsection{Active learning} \label{low.resource.active}

Active learning is a form of semi-supervised learning in which the learning algorithm can select the data from which it wants to learn, potentially improving its performance with respect to traditional learning approaches. One of the primary challenges in active learning is determining which data points are the most informative. The most widely used strategy today is uncertainty sampling \citep{settles2009active}, where the model selects examples for which its current predictions are the least certain. This strategy is effective because it focuses the learning on the most ambiguous examples, which helps refine the model's understanding.

When active learning is applied successfully to NLP, it can either improve model performance with the same amount of labeled data, or alternatively maintain similar performance while reducing the amount of data and annotation necessary. In deep learning research, active learning techniques have shown promising results. For example, \citet{siddhant2018deep} explored the use of uncertainty-based strategies such as Least Confident (LC) and Monte Carlo Dropout Bayesian Active Learning (DO-BALD), demonstrating the ability of these techniques to reduce the annotation requirement in deep learning models. \citet{shen2017deep} introduced Maximum Normalized Log-Probability (MNLP), an improvement over LC that normalizes uncertainty scores by sequence length, which is particularly effective in sequence-labeling tasks like NER.

In the context of NER, several deep learning approaches have adopted active learning strategies to improve model performance with limited annotations. For example, \citep{yan2023deep} investigated approaches that combine uncertainty- and diversity-based sampling to efficiently select the most informative examples for sequence labeling. Their work illustrates how gradient embeddings and clustering techniques, such as weighted k-means++, can be used to achieve a balance between informative and diverse sample selection, thus enhancing learning efficiency in deep active learning setups.

\subsection{Few-shot learning} \label{low.resource.few.shot}

Few-shot learning aims to build accurate machine learning models with minimal training data. This technique can be implemented by applying transformations to the data, applying changes to the algorithms, or using dedicated algorithms \citep{wang2020generalizing}. Applying transformations to the data involves generating new data from the training data using data augmentation or a generative network. Changes in the algorithms involve using pre-trained models as feature extractors or refining already trained models with new data through continued backpropagation. Dedicated algorithms involve networks that learn from pairs or triplets of instances rather than single instances, thereby leveraging a larger training set.

In the context of named entities, studies such as \citet{fritzler2019few} and \citet{hou2020few} have proposed the adaptation of prototypical networks \citep{snell2017prototypical} for NER. However, these implementations were unable to achieve optimal performance. \citet{yang2020simple} introduced a few-shot learning method based on nearest neighbors and structured inference that is shown to be superior to classical meta-learning approaches. \citet{cui2021template} approached the NER task as a language template classification problem, outperforming traditional sequence-labeling methods. An increasing number of works are recognizing the potential of few-shot learning in NER \citep{hofer2018few, huang2020few, he2023template}.

The emergence of LLMs, such as those utilized in PromptNER \citep{ashok2023promptner}, has further advanced few-shot learning in NER. These LLMs employ prompt-based methods and Chain-of-Thought prompting, significantly enhancing adaptability and performance in few-shot settings without extensive dataset requirements.

In this context, the CoTea framework \citep{yang2024cotea} represents a novel approach for low-resource NER, utilizing a divide-and-conquer strategy with two collaborative teacher models to improve training with minimal labeled data. CoTea leverages external knowledge and employs a mining refinery mechanism to iteratively improve label quality, thereby reducing noise and increasing performance, achieving competitive results even in extremely low-resource settings.

\subsection{Zero-shot learning} \label{low.resource.zero.shot}

Zero-shot learning uses a pre-trained model to assign classes to elements that the model has never encountered previously \citep{larochelle2008zero, lampert2013attribute, ding2017low}. This approach has been explored for linking entities \citep{wu2020scalable} and typing named entities \citep{obeidat2019description} (i.e., attributing a semantic label to a given entity).

Zero-shot learning can be applied in NER to detect new types of named entities. \citet{aly2021leveraging} proposed an architecture using textual descriptions. The ZERO model \citep{van2021zero} performs both zero-shot and few-shot learning by incorporating external knowledge through semantic representations of words. \citet{yang2022crop} proposed multilingual sequence translation as a solution for low-resource languages, where labeled data is scarce or absent. This method acts as a bridge by transferring knowledge from a source language to a target language with ample annotated data. Furthermore, the rise of prompt-based learning methods, as detailed in \citep{de2022entities}, has introduced a new paradigm in training and fine-tuning LLMs for applications like NER, enhancing the capabilities of zero-shot learning in this area. In a recent study \citep{gonzalez2023yes}, ChatGPT was evaluated for its zero-shot capabilities in NER on historical documents. The study found that, while ChatGPT has some capacity to identify entities, it struggles to cope with difficulties such as inconsistencies in annotation guidelines, complex entities, code-switching, and the accessibility of historical archives. Other references are explained in Section 5.5.
\section{Software frameworks} \label{frameworks}
NER has evolved significantly, and today a number of software frameworks provide robust tools to build and deploy NER systems. These frameworks simplify the development process by offering pre-built models, customizable pipelines, and support for various languages and domains. In this section, we present some of the most widely recognized and commonly used NER frameworks, highlighting their key features and capabilities.
\begin{itemize}
    \item \textbf{OpenAI} \citep{OpenAI2024} offers a range of AI tools, including GPT models, for text generation, question answering, and more. Although not originally focused on NER, OpenAI models can be adapted for NER tasks through fine-tuning or prompt engineering. The API is known for its flexibility and user-friendliness, with an additional emphasis on safe, ethical AI use.
    \item \textbf{spaCy} \citep{spacy2} is a free open-source library for advanced NLP in Python. It is designed to make it easy to construct systems for information extraction or general-purpose NLP. spaCy offers multiple analysis tools, such as tokenization, classification, POS tagging, and NER. In addition to the entities included by default, spaCy allows the addition of new classes by training the models on new data. A variety of pre-trained models are available, which can either be used directly for tasks such as NER or re-trained on specific datasets. These models are based on CNNs or Transformers.
    \item \textbf{NLTK} \citep{bird2006nltk} is a suite of Python modules for NLP, integrating more than 50 corpora and lexical resources such as WordNet, as well as a suite of tools for text analysis, including tokenization, POS tagging, sentiment analysis, topic segmentation, and speech recognition. Unlike spaCy, which includes built-in algorithms tailored to various tasks, NLTK provides flexibility by allowing users to choose among a wide range of algorithms.
    \item \textbf{Stanford CoreNLP} \citep{manning2014stanford} is a library developed by the associated research group at Stanford University. It is a set of natural language analysis tools written in Java, supporting tokenization, POS tagging, and training models for NER (based on CRF). NER features are only available for specific languages: English, Spanish, German, and Chinese, as each language has its unique set of characteristics.
    \item \textbf{Apache OpenNLP} \citep{baldridge2005opennlp} is a library that supports common NLP tasks such as NER, language detection, POS tagging, and chunking. Unlike other frameworks, which may use a single model for all entity types, OpenNLP provides a specialized maximum entropy model for each named entity type.
    \item \textbf{Polyglot} \citep{al2015polyglot} is an NLP pipeline for Python. It can handle a much wider range of languages than other frameworks, supporting NER in over 40 languages.
    \item \textbf{Flair} \citep{akbik2019flair} is a free, open-source library that enables the creation of NLP pipelines for multilingual applications. Flair allows the stacking of embeddings, meaning users can combine different embeddings (such as Flair, ELMo, and BERT) to improve NER performance. It supports various language models, including Flair embeddings, ELMo, and BERT.
    \item \textbf{Hugging Face} \citep{wolf2020Transformers} provides open-source NLP technologies. It offers both free and paid services aimed at businesses. The framework is particularly known for its Transformers library, which offers an API for accessing numerous pre-trained models, as well as the Datasets library, which simplifies managing datasets for NLP tasks. Hugging Face also includes a collaborative platform where users can create, train, and share their deep learning models.
    \item \textbf{Gate} \citep{cunningham2002gate} is a tool written in Java. It is used by a number of NLP communities for different languages. Gate provides an information extraction system, known as ANNIE, which is able to recognize several types of entities (people, places, and organizations).
    \item \textbf{TNER} \citep{ushio2022t} is a Python library for training and tuning NER models implemented in PyTorch. It features a web application with an intuitive interface that allows users to visualize predictions.
    \item \textbf{GliNER} \citep{zaratiana2023gliner} is a specialized NER model that leverages bidirectional Transformers, such as DeBERTa, for NER. Unlike traditional autoregressive models, GliNER supports parallel processing of entity spans, making it more efficient in resource-limited scenarios. It is designed to identify a wide variety of entity types by matching entity embeddings with text spans in a shared latent space.
\end{itemize}

Packages like Apache OpenNLP, Stanford CoreNLP, and spaCy are also accessible in languages other than Python. For example, openNLP\footnote{https://cran.r-project.org/web/packages/openNLP/index.html} is an R package that takes advantage of the capabilities of the Apache OpenNLP library, originally Java-based, by acting as an interface within the R environment. Similarly, the spacyr\footnote{https://cran.r-project.org/web/packages/spacyr/index.html} package connects R to spaCy. Notably, the spacyr package facilitates NER using spaCy's pre-trained language models. Other solutions like the reticulate\footnote{https://cran.r-project.org/web/packages/reticulate/index.html} package make it easier to achieve interoperability between R and Python, enabling Python libraries such as Hugging Face to be accessible within R.

\section{Evaluation of NER systems} \label{evaluation}
Evaluation of NER systems requires an annotation scheme, an evaluation strategy, and metrics. Each of these requirements is discussed below.

\subsection{Evaluation strategies: exact or relaxed evaluation} \label{evaluation.ner}
The evaluation of NER systems is based on comparing predictions with a gold standard and typically employs one of two strategies: exact evaluation or relaxed evaluation.
\begin{itemize}
    \item \textbf{Exact Evaluation}: In this approach, both the boundaries and the type of the named entity must match the gold standard accurately. This stringent method requires a perfect alignment between the predicted entity and the reference and is commonly used in the CoNLL-2003 evaluation \citep{0306050}.
    \item \textbf{Relaxed Evaluation}: This approach allows partial credit, giving points when either the type or the boundaries are correct, even if both are not. Relaxed evaluation is often used in the MUC \citep{grishman1996message} and ACE \citep{doddington2004automatic} standards.
\end{itemize}


\subsection{Metrics}\label{metrics}
Classical metrics such as precision, recall, and F1 score are often used for evaluating named entities:
\begin{itemize}
\item \textbf{Precision}: The proportion of named entities correctly recognized by the model in relation to the total number of named entities recognized. This metric reflects how many of the entities identified by the model are actually correct.
\begin{equation}
\text{precision} = \frac{\text{TP}}{\text{TP + FP}}
\end{equation}
where TP is the number of True Positives and FP is the number of False Positives.

\item \textbf{Recall}: The proportion of relevant named entities correctly retrieved by the model in comparison to the total number of relevant named entities in the dataset. This metric indicates how many of the actual entities the model successfully identified.
\begin{equation}
\text{recall} = \frac{\text{TP}}{\text{TP + FN}}
\end{equation}
where FN is the number of False Negatives.

\item \textbf{F1 score}: Reflects a model's effectiveness in detecting named entities by balancing precision and recall. It is calculated as the harmonic mean of precision and recall, providing a single measure that combines both metrics.
\begin{equation}
\text{F1} = 2 \times \frac{\text{precision} \times \text{recall}}{\text{precision} + \text{recall}}
\end{equation}

\end{itemize}
These metrics can be computed for each class of entities and can be aggregated when considering more than one type of entity:
\begin{itemize}
    \item \textbf{Macro-average}: The metric (e.g., F1 score) is computed for each class separately, and the macro-average is the mean of these values. This approach treats all classes equally, regardless of their frequency in the dataset.
    \item \textbf{Micro-average}: This method gives equal weight to each individual sample by pooling all predictions across classes before calculating the metrics.
\end{itemize}

To go beyond these aggregated metrics, \citep{fu-etal-2020-interpretable} proposed a new evaluation method involving a set of attributes possessed by entities (such as length or density). They found that models often have a better correlation with some attributes than with others, providing deeper insight into model performance.

\section{NER datasets} \label{datasets}
Named entities often belong to broad categories, such as persons, locations, and organizations. However, categories can be much narrower than this: for example, they might correspond to books, periodicals, magazines, etc. Table \ref{datasets} provides an overview of several English NER datasets, with between one and 505 types of entities in various domains such as medical data, news, social media, and more.

\begin{table}[ht]
\centering
\scriptsize
\caption{\label{citation-guide}
Datasets for English NER. Datasets highlighted in gray are those selected for our study.}
\begin{tabular}{ccccc}
\hline
Dataset & Year & Domain & Tags & URL\\
\hline
MUC-6 & 1995 & News & 7 & \url{https://cs.nyu.edu/~grishman/muc6.html} \\
MUC-7 & 1997 & News & 7 & \url{https://catalog.ldc.upenn.edu/LDC2001T02} \\
NIST-IEER & 1999 & News & 3 & \url{https://www.nist.gov/el/intelligent-systems-division-73500/ieee-1588} \\
CoNLL-2002 & 2002 & News & 4 & \url{https://www.clips.uantwerpen.be/conll2002/ner/}` \\
\rowcolor{Gray}
CoNLL-2003 & 2003 & News & 4 &
\url{https://www.clips.uantwerpen.be/conll2003/ner/} \\
GENIA & 2003 & Medical & 5 & \url{http://www.geniaproject.org/genia-corpus} \\
\rowcolor{Gray}
NCBI Disease & 2014 & Medical & 1 & \url{https://www.ncbi.nlm.nih.gov/pmc/articles/PMC3951655/} \\
i2b2-2014 & 2015 & Medical & 32 & \url{https://www.i2b2.org/NLP/DataSets/Main.php} \\
\rowcolor{Gray}
BC5CDR & 2016 & Medical & 2 & \url{https://biocreative.bioinformatics.udel.edu/tasks/biocreative-v/track-3-cdr/} \\
MedMentions & 2019 & Medical & 128 & \url{https://github.com/chanzuckerberg/MedMentions} \\
\rowcolor{Gray}
BioNLP2004 & 2004 & Bioinformatics & 5 & \url{https://www.ncbi.nlm.nih.gov/research/bionlp/Data/} \\
ACE 2004 & 2005 & Various & 7 & \url{https://catalog.ldc.upenn.edu/LDC2005T09} \\
ACE 2005 & 2006 & Various & 7 & \url{https://catalog.ldc.upenn.edu/LDC2006T06} \\
\rowcolor{Gray}
OntoNotes 5.0 & 2013 & Various & 18 & \url{https://catalog.ldc.upenn.edu/LDC2013T19} \\
\rowcolor{Gray}
MultiCoNER & 2022 & Various & 33 & \url{https://multiconer.github.io/} \\
WikiGold & 2009 & Wikipedia & 4 & \url{https://aclanthology.org/W09-3302} \\
WiNER & 2012 & Wikipedia & 4 & \url{https://github.com/ghaddarAbs/WiNER} \\
WikiFiger & 2012 & Wikipedia & 112 & \url{https://orkg.org/paper/R163134} \\
\rowcolor{Gray}
Few-NERD & 2021 & Wikipedia & 66 & \url{https://github.com/thunlp/Few-NERD} \\
HYENA & 2012 & Wikipedia & 505 & \url{https://aclanthology.org/C12-2133.pdf} \\
WikiAnn & 2017 & Wikipedia & 3 & \url{https://aclanthology.org/P17-1178/} \\
\rowcolor{Gray}
WNUT 2017 & 2017 & Social media	& 6 & \url{https://noisy-text.github.io/2017/emerging-rare-entities.html} \\
MalwareTextDB & 2017 & Malware & 4 & \url{https://statnlp-research.github.io/resources/} \\
SciERC & 2018 & Scientific & 6 & \url{http://nlp.cs.washington.edu/sciIE/} \\
HIPE-2022-data & 2022 & Historical & 3 & \url{https://github.com/hipe-eval/HIPE-2022-data} \\
MITMovie & 2013 & Queries & 12 & \url{http://groups.csail.mit.edu/sls/} \\
\rowcolor{Gray}
MITRestaurant & 2013 & Queries & 8 & \url{http://groups.csail.mit.edu/sls/} \\
\rowcolor{Gray}
FIN & 2015 & Financial & 4 & \url{https://aclanthology.org/U15-1010/} \\
\hline
\end{tabular}
\end{table}

In the remainder of our survey and in our experiments, we make use of the following datasets obtained from various sources:

\begin{itemize}
    \item \textbf{CoNLL-2003}: This dataset consists mainly of news articles from Reuters.
    \item \textbf{OntoNotes 5.0}: A comprehensive dataset comprising various genres of texts including phone conversations, newswires, newsgroups, broadcast news, broadcast conversations, weblogs, and religious texts.
    \item \textbf{WNUT2017}: This dataset includes texts from various sources, such as tweets, Reddit comments, YouTube comments, and StackExchange.
    \item \textbf{BioNLP2004}: A biomedical dataset comprising 2000 abstracts from the MEDLINE database, annotated for NER.
    \item \textbf{FIN}: A dataset containing financial documents released by the US Securities and Exchange Commission, annotated specifically for financial named entities \citep{alvarado2015domain}.
    \item \textbf{NCBI Disease}: This dataset provides disease names and concept annotations drawn from the NCBI Disease Corpus, with a focus on biomedical-named entities.
    \item \textbf{BC5CDR}: A dataset consisting of articles with annotations for chemicals, diseases, and their relationships.
    \item \textbf{MITRestaurant}: A collection of annotated online restaurant reviews for entity recognition in the culinary domain.
    \item \textbf{Few-NERD}: This dataset contains a collection of Wikipedia articles and news reports. Due to its large size, we trained the models on a limited portion (20\% of the data, which represents 32,941 samples).
    \item \textbf{MultiCoNER}: A large multilingual dataset covering three domains: Wiki sentences, questions, and search queries.

\end{itemize}
The characteristics of these datasets are provided in Table \ref{chosen.datasets}.

\begin{table}[!ht]
\small
\centering
\caption{Characteristics of the selected datasets used in our comparative study, with “\#” indicating the number of samples in each split.}
\begin{tabular}{ccccc}
\hline
{\bf Corpus} & {\bf \#Train} & {\bf \#Test} & {\bf \#Validation} & {\bf tags}\\
\hline
CoNLL-2003 & 14,041 & 3,453 & 3,250 & 4 \\
OntoNotes 5.0 & 59,924 & 8,262 & 8,528 & 18 \\
WNUT2017 & 2,395 & 1,287 & 1,009 & 6 \\
BioNLP2004 & 16,619 & 3,856 & 1,927 & 5 \\
FIN & 1,018 & 305 & 150 & 4 \\
NCBI Disease & 5,433 & 941 & 924 & 1 \\
BC5CDR & 5,228 & 5,865 & 5,330 & 2 \\
MITRestaurant & 6,900 & 1,521 & 760 & 8 \\
Few-NERD & 131,767 & 37,648 & 18,824 & 66 \\
MultiCoNER & 16,778 & 249,980 & 871 & 33 \\
\hline
\end{tabular}
\label{chosen.datasets}
\end{table}
\section{Experiments}\label{sec:experiments}

In this section, we describe the methodology used to assess the performance of the chosen algorithms.

\subsection{Datasets} Our experiments were carried out on ten datasets from various domains, as shown in Table \ref{chosen.datasets}. These datasets differ in size and class count, allowing for various evaluation scenarios.

The corpora were formatted using the CoNLL-U standard, following the BIO tagging scheme, where (a) each word line includes annotations for individual words, and (b) blank lines denote sentence boundaries. Since each framework requires its own data representation format, we converted the original CoNLL-U format into the appropriate formats for each framework.

For GPT-4, custom prompts were designed for each dataset to highlight the key categories of named entities specific to that dataset. These prompts also included multiple examples to provide context. The box below showcases sample prompts for the BIONLP2004 dataset.

\begin{tcolorbox}[colback=blue!5!white, colframe=blue!75!black, title=Example of a prompt used for the dataset BIONLP2004]
\label{box:example_prompt}
\scriptsize
Identify the named entities in the following sentence, categorizing them according to these definitions:
\begin{itemize}
    \item \textbf{"DNA"}:
        \begin{itemize}
            \item Entities that refer to specific DNA sequences or are related to DNA in a biological context.
            \item Examples: BRCA1 gene, CD14 5'-Upstream Sequence, or nonoptimal binding sites.
        \end{itemize}
    \item \textbf{"Protein"}:
        \begin{itemize}
            \item Entities representing specific proteins or related to proteins in a biological sense.
            \item Examples: c-myb, hemoglobin, or E-box-binding repressor.
        \end{itemize}
    \item \textbf{"Cell\_type"}:
        \begin{itemize}
            \item Entities referring to specific types of cells in a biological context.
            \item Examples: neuron, T-cell, or hepatocyte.
        \end{itemize}
    \item \textbf{"Cell\_line"}:
        \begin{itemize}
            \item Entities representing specific cell lines used in biological research.
            \item Examples: Monocytic U937 cells, Jurkat cell line, or MCF-7.
        \end{itemize}
    \item \textbf{"RNA"}:
        \begin{itemize}
            \item Entities that are specific RNA sequences or related to RNA in a biological context.
            \item Examples: mRNA, siRNA, or c-jun mRNA.
        \end{itemize}
\end{itemize}

Format the output in JSON with keys corresponding to each entity type. List the identified named entities under each key, grouping multiple entities of the same type into a single list. Ensure a clear separation and precise identification of each entity to avoid ambiguity. Return only the JSON response with no additional explanations.
\end{tcolorbox}

In our study, we adopted an experimental approach that diverges slightly from the method proposed in the GPT-NER paper \cite{wang2023gptner}. While that study demonstrated an effective technique, it relied on distinct prompts for each category of named entities, querying the model individually. This approach, though successful, simplifies the model's task by reducing the need to disambiguate between different entity types. To better simulate real-world conditions, we opted for a less guided strategy that challenges the model's general capability to accurately identify and classify named entities in more complex and varied scenarios. This approach enhances the evaluation by requiring greater context understanding and entity disambiguation, which are essential to the model's overall effectiveness in practical applications.

\subsection{Models}
Our selection of frameworks was guided by three primary criteria: open-source availability, free access, and the ability to train models on custom datasets. Based on these factors, we included Flair \cite{akbik2019flair}, Stanford CoreNLP \cite{manning2014stanford}, spaCy \cite{spacy2}, GliNER \cite{zaratiana2023gliner}, OpenAI \cite{OpenAI2024}, and Hugging Face \cite{wolf2020Transformers} in our analysis. For frameworks that support multiple algorithms, we selected several representative models.

For spaCy, we evaluated three models, each illustrating different architectures and capability levels: a small CNN (``\texttt{en\_core\_web\_sm}''), a large CNN (``\texttt{en\_core\_web\_lg}''), and a Transformer model based on basic RoBERTa \cite{liu2019roberta} (``\texttt{en\_core\_web\_trf}''). 

In the case of Hugging Face, we chose six models: the basic BERT \cite{devlin2018bert} architecture in both lowercase and uppercase configurations (``\texttt{BERT-base-cased}'' and ``\texttt{BERT-base-uncased}''), its distilled variant \cite{sanh2020distilbert} (``\texttt{DistilBERT-base-cased}''), a large RoBERTa-based model (``\texttt{FacebookAI/xlm-roberta-large}''), and both small and large versions of DeBERTa \cite{he2020deberta} (``\texttt{Microsoft/DeBERTa-v3-base}'' and ``\texttt{Microsoft/DeBERTa-v3-large}''). 

For OpenAI, we used the GPT-4o\footnote{https://openai.com/index/hello-gpt-4o/} model, while for GliNER, we selected the large architecture based on DeBERTa (``\texttt{urchade/gliner\_large}''), both in its pre-trained and fine-tuned versions.

Turning to Flair, we leveraged the standard architecture, which combines an LSTM-CRF network with Flair embeddings. A grid search determined the optimal hyperparameters, testing hidden layer sizes of 64, 128, and 256, and learning rates of 0.05, 0.1, 0.15, and 0.2. The chosen configuration, which achieved the highest F1 score on the validation set, used a learning rate of 0.1 and a hidden layer size of 256.

For the Hugging Face models, we used the TNER library \cite{ushio2022t} to fine-tune each model, with entity decoding managed by CRF \cite{lafferty2001conditional}. Hyperparameter tuning involved a grid search with learning rates of $10^{-4}$ and $10^{-5}$, both with and without weight decay set to 0.01. Training was carried out over 10 epochs, with an interim evaluation at 5 epochs, a batch size of 16, and gradient accumulation steps set to 2. Additionally, configurations included both CRF and non-CRF models, with a warmup step ratio of 0.1.

For both Stanford CoreNLP and spaCy, we applied their default settings to maintain consistency with standard configurations.

Lastly, for GliNER, we fine-tuned the large model (``\texttt{urchade/gliner\_large}'') on each dataset using customized hyperparameters. The main model was trained with a learning rate of $5 \times 10^{-6}$ and weight decay set to 0.01. We used a linear learning rate scheduler with a warm-up ratio of 0.1. Training was carried out over 10 epochs with a batch size of 16, and focal loss parameters were set to alpha = 0.75 and gamma = 2. Step-based evaluations were performed every 300 steps.

\subsection{Evaluation and statistical analysis}
To evaluate the results, we use an exact evaluation. We select the F1 score as our primary metric because it includes the other two metrics mentioned in Section \ref{metrics}, specifically precision and recall.

To assess the differences between the models, we perform the Friedman test \citep{friedman1940comparison}. The Friedman test is a non-parametric statistical test used to detect differences in treatments across multiple attempts. The models are ranked according to their F1 scores, and the null hypothesis of the Friedman test is that the medians of these rankings are equal across the models.

The Friedman test ranks each block together and compares the rankings between columns. The test statistic is given by:
\begin{equation}
    Q = \frac{12}{nk(k+1)} \sum_{j=1}^{k} R_j^2 - 3n(k+1)
\end{equation}
where:
\begin{itemize}
\item $n$ is the number of blocks (different sets of data being compared),
\item $k$ is the number of models,
\item $R_j$ is the sum of the rankings for the $j$-th treatment.
\end{itemize}

When the null hypothesis of the Friedman test is rejected (given a chosen significance threshold, e.g., $\alpha = 0.1$), we apply Nemenyi's method \citep{nemenyi1963distribution} to identify the model pairs that differ significantly. Nemenyi's post-hoc test compares the average rankings of executions to determine if the differences are statistically significant. The critical difference (CD) is computed as follows, to determine if the differences in average rankings between model pairs are statistically significant: \begin{equation} CD = q_\alpha \sqrt{\frac{k(k+1)}{6n}} \end{equation} where $q_\alpha$ is the critical value from the studentized range distribution, based on the selected significance level $\alpha$.
\section{Results and discussion}
Table \ref{table:results} provides a comparison of the selected NER frameworks based on their macro-averaged F1 scores across ten datasets. The best and second-best performances are highlighted in bold and underlined, respectively. Our experiments reveal key performance trends influenced by dataset size, domain specificity, and the comparative effectiveness of pre-trained versus fine-tuned NER models. In the following, we delve into the details of these results and their implications for model performance across various datasets.

\begin{table}[h!]
\small
\label{tab.res.exp}
\caption{Comparison of NER frameworks in terms of Macro-averaged F1 score. Best and second-best scores are respectively in bold and underlined. }
\begin{adjustbox}{width=\textwidth}
\begin{tabular}{cccccccccccc}
\cline{3-12}
 \multicolumn{2}{c}{}  &
 \rot{CoNLL-2003} & \rot{OntoNotes} & \rot{WNUT2017} & \rot{FIN} & \rot{BioNLP2004} & \rot{NCBI Disease} & \rot{BC5CDR} & \rot{MITRestaurant} & \rot{Few-NERD} & \rot{MultiCoNER} \\ 
 \toprule 
\textbf{Frameworks} & \textbf{Algorithms} & \multicolumn{10}{c}{\textbf{Macro-averaged F1 score}} \\ 
 \midrule 
Stanford CoreNLP & CRF &  86.96 & 66.61 & 13.18 & 56.70 & \textbf{79.95} & \textbf{90.94} & 88.80 & 74.35 & 47.07 & 20.00 
\\ \midrule 
 Flair & LSTM-CRF & 90.35 & 80.08 & 38.07 & \textbf{69.22} & 71.64 & 85.97 & \textbf{90.51} & 79.19 & 60.28 & 55.81 \\ \midrule 

 GliNER & GliNER-L pre-trained &  48.73 & 28.41 & 36.77 & 15.59 & 45.53 & 64.16 & 68.65 & 35.09 & 34.63 & 27.73 \\
 & GliNER-L fine-tuned &  90.70 & \textbf{83.15} & \textbf{54.38} & \underline{65.46} & \underline{73.13} & \underline{89.23} & \underline{90.10} & \underline{80.42} & \textbf{64.96} & \underline{63.19} \\ \midrule 
 spaCy & en\_core\_web\_sm &  80.54 & 68.56 & 9.72 & 54.08 & 66.85 & 78.71 & 80.28 & 74.86 & 39.27 & 36.82 \\
  & en\_core\_web\_lg & 81.66 & 69.68 & 9.57 & 56.52 & 65.95 & 78.54 & 80.25 & 75.57 & 40.20 & 35.36 \\
 & en\_core\_web\_trf & 90.36 & 80.79 & 39.35 & 53.43 & 71.34 & 86.68 & 87.26 & 78.48 & 60.81 & \textbf{63.82} \\ 
\midrule 
Hugging Face & FacebookAI/xlm-RoBERTa-large & 90.85 & 80.63 & 41.04 & 46.33 & 70.35 & 86.49 & 87.91 & \underline{80.45} & 61.05 & 58.68 \\
 & DistilBERT-base-cased & 88.08 & 77.66 & 24.11 & 42.08 & 67.25 & 84.41 & 83.99 & 77.85 & 57.66 & 54.34 \\
 & BERT-base-uncased & 88.87 & 77.64 & 33.78 & 39.67 & 69.31 & 85.89 & 85.06 & 79.44 & 58.29 & 59.99 \\
 & BERT-base-cased & 89.37 & 79.52 & 33.69 & 38.43 & 67.60 & 86.59 & 85.21 & 77.32 & 59.56 & 55.88 \\
 & Microsoft/DeBERTa-v3-base & \textbf{91.36} & 80.13 & 42.96 & 46.94 & 71.21 & 88.03 & 89.20 & 79.47 & 60.44 & 56.46 \\
 & Microsoft/DeBERTa-v3-large & \underline{90.98} & \underline{82.66} & \underline{44.81} & 47.28 & 71.61 & 86.70 & 89.20 & \textbf{81.06} & \underline{61.83} & 61.94 \\ 
\midrule 
OpenAI & GPT-4o & 65.17 & 59.12 & 44.73 & 36.70 & 45.32 & 63.46 & 67.62 & 51.72 & 50.87 & 40.90\\
\bottomrule
\end{tabular}
\end{adjustbox}
\label{table:results}
\end{table}

Starting with large general-domain datasets such as CoNLL-2003, OntoNotes, Few-NERD, and MultiCoNER, Transformer-based models such as GliNER-L (fine-tuned), RoBERTa, and DeBERTa demonstrate outstanding performance. This can probably be attributed to their high parameter counts, which enable them to capture complex data relationships when trained on extensive datasets. A key insight from this analysis is the notable impact of GliNER-L's comprehensive pre-training and fine-tuning processes. Compared to DeBERTa from Microsoft, GliNER-L exhibits improved performance, especially on domain-specific datasets. This improvement is likely due to GliNER-L's structured approach, which involves sequential stages: pre-training as a language model, followed by NER training, and fine-tuning on domain-specific datasets. These stages allow GliNER-L to capture nuanced entity relationships and effectively adapt to NER tasks.

Shifting the focus to domain-specific datasets, particularly in the biomedical field (BioNLP2004, NCBI Disease, and BC5CDR), a different trend emerges. Traditional models such as CRF and LSTM-CRF remain competitive, often closely matching or even surpassing Transformer-based models. This may be due to the specialized terminology in biomedical texts, which is not well represented in the general datasets typically used for Transformer pre-training. However, fine-tuning on domain-specific data, as demonstrated with GliNER-L, significantly enhances performance by refining embeddings to better capture the contextual nuances of specialized terms. This trend is illustrated in Table \ref{result.bio}, which compares the performance of CRF, LSTM-CRF, GliNER-L (fine-tuned) and Microsoft's DeBERTa on key types of entity in biomedical datasets.

\begin{table}[h!]
\centering
\small
\caption{Performance comparison of top models on biomedical datasets (BC5CDR, BioNLP2004, NCBI Disease) for key entity types, showing F1, precision, and recall scores for each model}
\setlength{\tabcolsep}{2pt} 
\begin{adjustbox}{width=\textwidth}
\begin{tabular}{llcccc@{\hskip 5pt}cccc@{\hskip 5pt}cccc@{\hskip 5pt}ccc}
\toprule
 & Model & \multicolumn{3}{c}{CRF} & & \multicolumn{3}{c}{LSTM-CRF} & & \multicolumn{3}{c}{GliNER-L} & & \multicolumn{3}{c}{DeBERTa (large)} \\ 
 & & F1 & Precision & Recall & & F1 & Precision & Recall & & F1 & Precision & Recall & & F1 & Precision & Recall \\
 \cline{3-5} \cline{7-9} \cline{11-13} \cline{15-17} 
Dataset & entity\_type &  &  &  &  &  &  &  &  &  &  &  &  \\
\midrule
\multirow{2}{*}{BC5CDR} & Chemical  & 89.83 & 99.0 & 82.22 & & 94.23 & 94.39 & 94.08 & & 93.63 & 93.44 & 93.83 & & 92.57 & 92.05 & 93.09 \\
 & Disease  & 87.77 & 99.41 & 78.56 & & 86.79 & 86.13 & 87.45 & & 86.56 & 84.56 & 88.65 & & 85.84 & 84.18 & 87.57 \\
\midrule
\multirow{5}{*}{BioNLP2004} & cell\_line  & 67.78 & 64.79 & 71.05 & & 60.83 & 54.52 & 68.8 & & 62.97 & 54.91 & 73.8 & & 62.25 & 53.61 & 74.2 \\
 & cell\_type  & 79.36 & 94.54 & 68.38 & & 76.3 & 81.7 & 71.58 & & 76.81 & 78.24 & 75.43 & & 76.12 & 80.1 & 72.51 \\
 & DNA  & 81.17 & 89.32 & 74.38 & & 73.4 & 72.23 & 74.62 & & 74.39 & 71.23 & 77.84 & & 73.61 & 70.51 & 76.99 \\
 & Protein  & 86.37 & 93.96 & 79.92 & & 76.67 & 71.01 & 83.32 & & 78.46 & 71.36 & 87.13 & & 77.19 & 70.83 & 84.8 \\
 & RNA  & 85.05 & 93.0 & 78.36 & & 70.97 & 67.69 & 74.58 & & 73.0 & 66.21 & 81.36 & & 68.86 & 60.65 & 79.66 \\
\midrule
NCBI Disease & Disease  & 90.94 & 99.9 & 83.39 & & 85.97 & 86.1 & 85.83 & & 89.23 & 86.73 & 91.88 & & 86.65 & 86.42 & 86.88 \\
\bottomrule
\end{tabular}
\end{adjustbox}
\label{result.bio}
\end{table}
In the BC5CDR dataset, which includes both ``Chemical'' and ``Disease'' entities, LSTM-CRF achieves the highest F1 score for the ``Chemical'' entity (94.23), outperforming both GliNER-L and DeBERTa. This suggests that LSTM-CRF may have an edge in identifying chemical entities, likely due to its structured feature handling. For the entity ``Disease'', CRF shows strong precision performance (99.41), but GliNER-L achieves the highest recall (88.65), indicating its ability to capture more disease-related entities even though its overall F1 score is slightly lower than LSTM-CRF.

In the BioNLP2004 dataset, which covers a broader range of entity types (``cell\_line'', ``cell\_type'', ``DNA'', ``Protein'', and ``RNA''), CRF consistently achieves high scores, especially in precision across entities like ``cell\_type'' (94.54), ``DNA'' (89.32), ``Protein'' (93.96), and ``RNA'' (93.0). For ``Protein'' and ``RNA'' entities, GliNER-L shows competitive recall values (87.13 and 81.36, respectively), indicating its strength in capturing complex biomedical terms despite not leading in all F1 scores. LSTM-CRF and DeBERTa display relatively balanced scores across categories, though GliNER-L has a slight edge in recall for multi-word biomedical terms.

In the NCBI Disease dataset, which focuses solely on the ``Disease'' entity, CRF leads with the highest F1 (90.94) and precision (99.9), underscoring its effectiveness in datasets with specialized terminology. However, GliNER-L achieves the highest recall (91.88), highlighting its ability to identify a broader range of disease-related mentions within the dataset.

Turning to smaller datasets such as FIN and WNUT2017, we observe that LSTM-CRF models tend to outperform Transformer-based models. This difference may stem from the tendency of overparameterized models, such as Transformers, to overfit when trained on limited data. For example, on the FIN dataset, LSTM-CRF achieves the highest F1 score (69.22), significantly outperforming Transformer models, which require careful hyperparameter tuning and well-structured validation sets to reduce overfitting. Notably, GliNER-L fine-tuned remains competitive even on these smaller datasets, consistently ranking among the top three models. This suggests that GliNER-L's structured pre-training and fine-tuning processes enhance its robustness and adaptability across datasets of varying sizes.

\begin{figure}[h!]
    \centering
    \includegraphics[width=12cm]{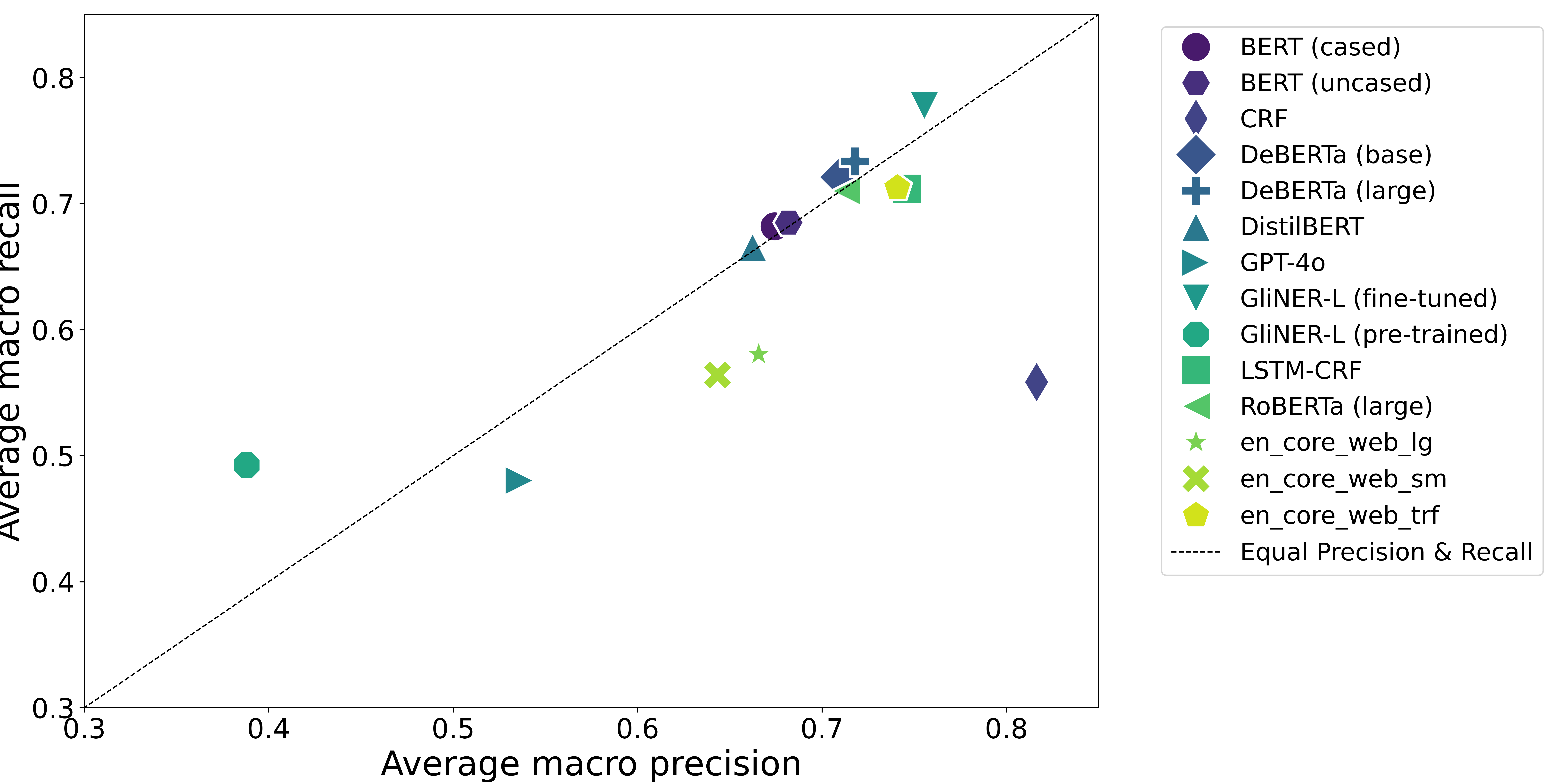}
    \caption{Average precision and recall scores by model. Each point represents a model's mean precision and recall across all datasets, with the dashed line indicating a balance between precision and recall.}
    \label{fig:precision_recall_plot}
\end{figure}

Figure \ref{fig:precision_recall_plot} represents the average scores of each model in all datasets. Each point shows the mean precision and recall for a model, with the dashed line indicating a perfect balance between the two metrics. Here are some insights:

\begin{itemize}
    \item \textbf{Precision/recall balance:} Models located near the central line, such as DeBERTa variants and LSTM-CRF, exhibit a well-balanced trade-off between precision and recall. This balance makes them particularly suitable for tasks where an equilibrium between these metrics is crucial, ensuring both high precision and coverage in entity detection.
    \item \textbf{Precision-focused models:} Models positioned above the reference line, such as CRF and CNNs (``\texttt{en\_core\_web\_sm}'' and ``\texttt{en\_core\_web\_lg}''), demonstrate higher precision than recall, making them ideal for applications where minimizing false positives is a priority.
    \item \textbf{Recall-focused model:} GliNER-L pre-trained model favors recall, demonstrating considerably higher recall than precision, though it still falls significantly behind the other models in overall performance.
\end{itemize}

Apart from GliNER-L pre-trained, which leans significantly toward recall, the other transformer models generally achieve a balanced approach, without a strong bias toward either precision or recall. GliNER-L pre-trained has allowed it to recognize a broad range of named entities, which boosts its recall by capturing many possible entities. However, this recall comes at the expense of precision, as the model often makes errors. To achieve effective performance, GliNER-L requires fine-tuning on specific datasets. In contrast, the CNN and CRF models appear to favor precision, likely due to their structural design and training objectives. For example, CRF models are optimized to model dependencies between labels in sequence data, enabling them to capture context and dependencies effectively. This makes CRFs more precise in identifying specific, well-defined entities, as they are less prone to false positives. Similarly, CNNs focus on capturing local patterns in the data, which can make them precise in NER tasks.

\begin{figure}[h!]
    \centering
    \includegraphics[width=\linewidth]{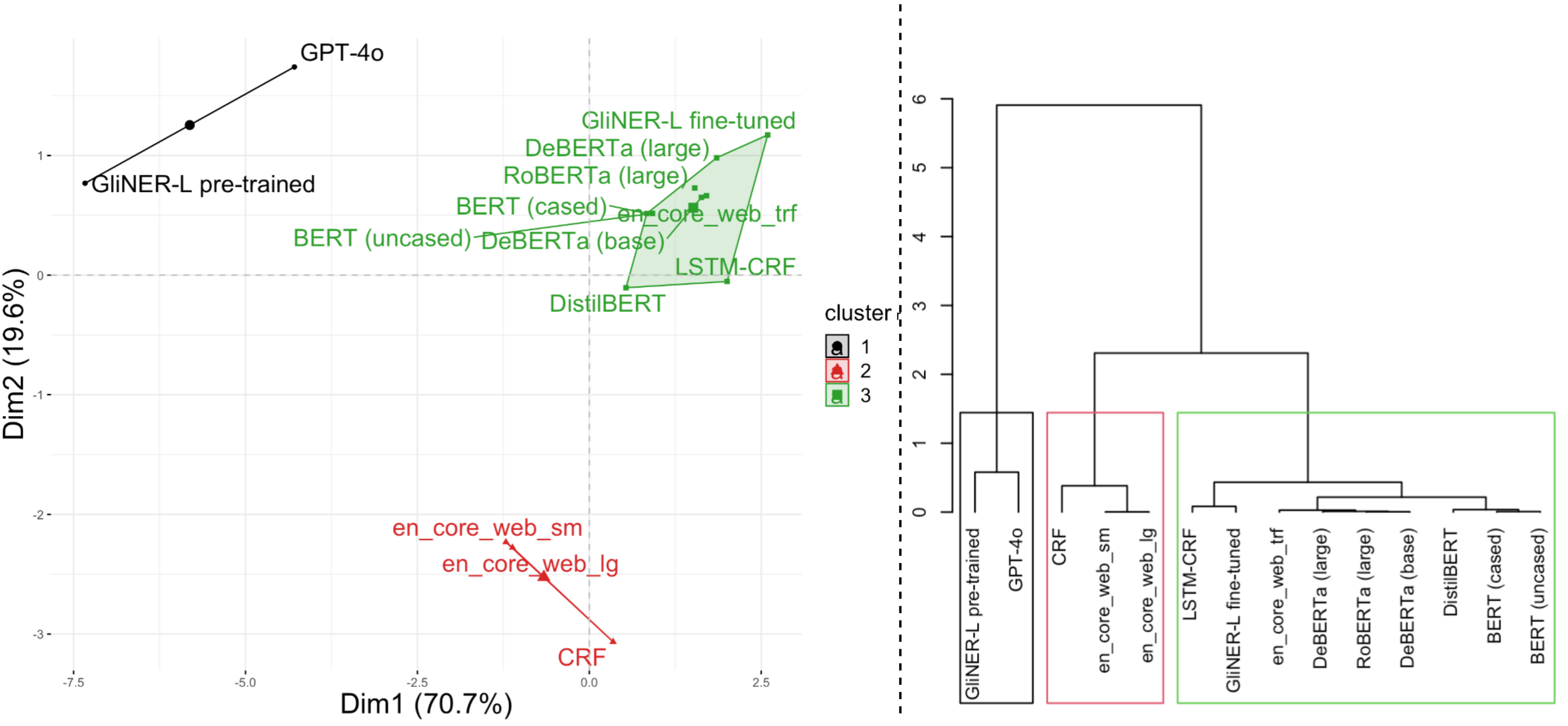}
    \caption{PCA plot (left) and hierarchical clustering dendrogram (right) of NER models based on macro-averaged F1 scores across all datasets.}
    \label{fig:pca_f1}
\end{figure}

Furthermore, Figure \ref{fig:pca_f1} presents a PCA plot (left) and hierarchical clustering (right), which illustrate performance similarities between frameworks. In the PCA plot, each point represents a model, with colors indicating clusters based on performance similarity. The first two dimensions capture most of the variance, showing that Transformer-based models, such as DeBERTa (base and large) and GliNER-L fine-tuned, cluster closely, suggesting consistent performance across datasets. In contrast, isolated points such as GPT-4o and GliNER-L pre-trained exhibit lower performance, particularly without fine-tuning. Hierarchical clustering (right) reinforces these insights. Traditional models, such as CRF and CNNs (both ``\texttt{en\_core\_web\_sm}'' and ``\texttt{en\_core\_web\_lg}''), form a distinct cluster, highlighting their competitiveness but limited flexibility compared to Transformers. Fine-tuned Transformers, especially GliNER-L, stand out, underscoring the advantages of domain-specific training.

\begin{figure}[!ht]
  \centering
\includegraphics[width=12cm]{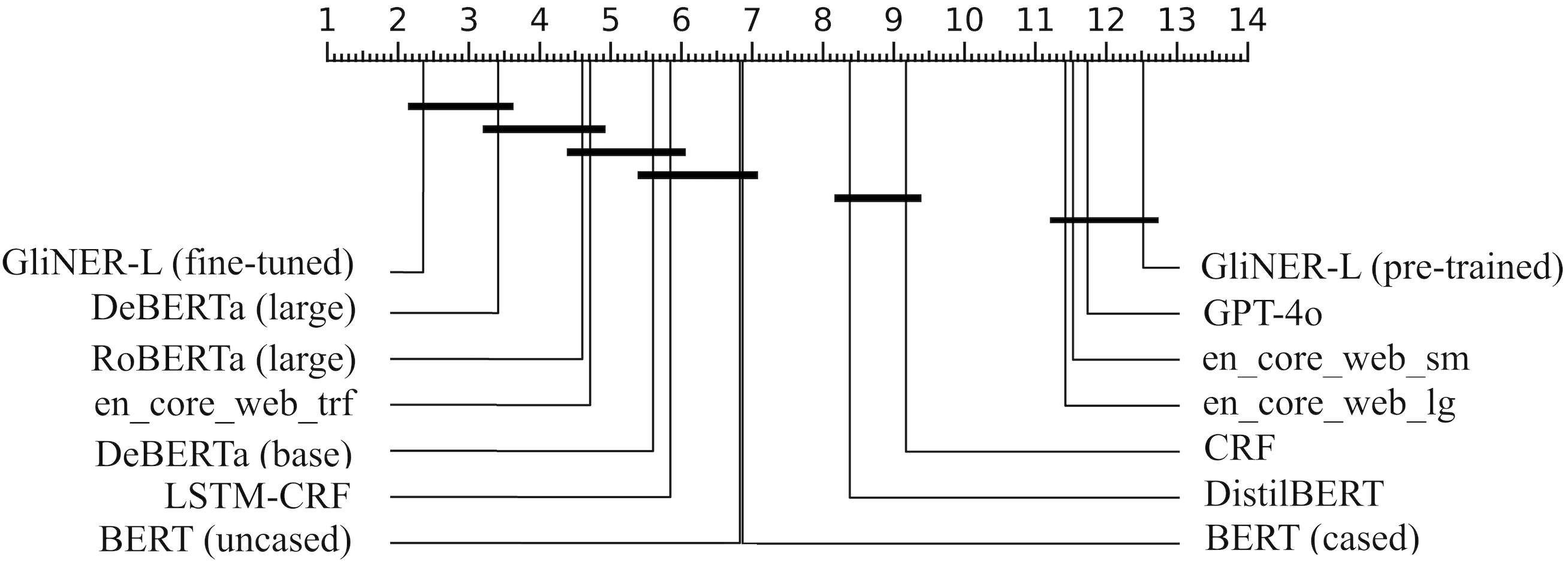}
  \caption{The critical difference (CD) diagram based on F1 score. The plot shows the mean rankings of the different models on 10 datasets. The lower the ranking, the better the performance of a model. A horizontal line indicates no significant difference between the models crossed by the line.}
  \label{nemenyi.test}
\end{figure}

Statistical analysis supports these findings. A Friedman test applied to all entity types in our datasets reveals statistically significant differences between model medians (p-value = 0.000), confirming the significance at an alpha level of 5\%. The results of Nemenyi's post hoc test (Figure \ref{nemenyi.test}) show that certain groups of models have similar performance levels, as indicated by the connecting horizontal lines. For example, models such as GliNER-L fine-tuned and DeBERTa (large) achieve the best rankings, indicating superior performance, and are not significantly different from each other. This suggests that these models perform similarly on the evaluated datasets. In particular, among the Transformer-based models, the DeBERTa variants appear to be the most effective for NER.

Additionally, another cluster of models, including GliNER-L pre-trained, CNNs (``\texttt{en\_core\_web\_sm}'' and ``\texttt{en\_core\_web\_lg}''), and GPT-4o, also exhibit statistically similar performance, although with higher mean ranks, indicating relatively lower performance compared to the top performing models. Toward the right of the figure, models such as CRF, DistilBERT, and BERT exhibit the highest ranks, indicating inferior performance in comparison to the other models.
Our experiments have shown that while LLMs perform exceptionally well in many NLP tasks, their general nature often leads to a lack of specialized effectiveness for NER, which can impede precise entity recognition, particularly in noisy or dynamic contexts. Evaluations highlight a performance gap for NER tasks \cite{han2023information}, likely due to limited task-specific learning and explicit understanding, which may lead to a ``lack of specialty'' in NER. Moreover, the large, sometimes undisclosed, number of parameters in LLMs poses challenges for fine-tuning to meet the demands of NER. Consequently, directly addressing NER tasks using only LLMs remains difficult. However, combining LLMs with simpler and fine-tuned NER models offers a promising solution. Approaches such as LinkNER \cite{zhang2024linkner} leverage a hybrid model in which the NER component handles common entities, while the LLM, guided by uncertainty estimation, addresses complex or ambiguous cases, enhancing robustness and flexibility in open domain contexts. Similarly, the Super In-Context Learning (SuperICL) approach \cite{xu2023small} integrates smaller specialized plug-in models to provide task-specific predictions and confidence scores, refining LLM outputs for more accurate final predictions. Together, LinkNER and SuperICL demonstrate that combining the broad contextual knowledge of LLMs with the focused precision of smaller models offers a promising path to advance NER, particularly in challenging open-domain settings with unseen entities and noisy data.

\section{Conclusion and perspectives} \label{conclusion}

This article presents a comprehensive survey on recent advances in NER within a classification framework. We focus on recent methods, including LLMs, graph-based approaches, reinforcement learning techniques, and strategies to train models on small datasets. To assess these approaches, we evaluated popular frameworks across datasets with varying characteristics.

Transformer-based architectures, especially on larger datasets, have demonstrated strong performance due to their substantial parameterization and adaptability. However, our analysis indicates that, despite the overall success of Transformers, models like GPT-4o do not consistently achieve top rankings for NER tasks. This may be attributed to the challenges in accurately disambiguating and detecting composite named entities, which are crucial in NER. In contrast, GliNER-L has shown remarkable consistency and robustness across various datasets, making it a reliable choice for a wide range of applications. Moreover, in the case of smaller datasets or specialized domains, such as biomedical texts, traditional approaches such as CRF or LSTM-CRF can outperform Transformers, as these simpler models often handle specific terminology and limited data more effectively.

Looking ahead, the potential of LLMs to enhance NER should not be underestimated. Despite current limitations in handling specialized NER tasks, the rapid advancement of language models presents an exciting opportunity to integrate these technologies into more refined NER systems. Promising directions to advance NER include hybrid approaches that combine LLMs with specialized, fine-tuned NER models (such as LinkNER or SuperICL). By blending the broad contextual understanding of LLMs with the precision of specialized NER models, these strategies hold promise to tackle NER in dynamic, open-domain settings that involve unseen entities and noisy data. Future research should further investigate these hybrid approaches, along with fine-tuning techniques and preprocessing methods designed to enhance LLM adaptability for NER. Such strategies could improve the applicability and accuracy of NER systems across diverse datasets and challenging contexts.

\bibliographystyle{unsrtnat}
\bibliography{references}

\end{document}